\documentclass[a4paper]{scrartcl}

 \usepackage{amsmath}
  \usepackage{amsthm}		 	    
 \usepackage{url}			
\usepackage{makeidx}
\makeindex

\usepackage{newtxtext}
 \usepackage[varbb]{newtxmath}
\usepackage[scaled=0.86]{helvet}
\usepackage{multirow}
\usepackage{graphicx}
\usepackage{hyperref}
\usepackage{subcaption}

\usepackage{algorithm}
\usepackage{algpseudocode}

\newcommand{\dx}{\mathrm{d}}

\newcommand{\R}{\mathbb{R}}

\newcommand{\E}{\mathbb{E}}

\newcommand{\tT}{\mathrm{T}}

\newcommand{\weakly}{\rightharpoonup}\normalfont

\DeclareMathOperator{\Lip}{Lip}
\newcommand*\samethanks[1][\value{footnote}]{\footnotemark[#1]}
\newcommand\addnote[1]{\captionsetup{font=small}\caption*{#1}}
\usepackage{mathtools}
\mathtoolsset{showonlyrefs}

\title{Generalized Normalizing Flows
via Markov Chains}

\author{Paul Lyonel Hagemann\thanks{TU Berlin,
Stra{\ss}e des 17. Juni 136, 
D-10623 Berlin, Germany,
\{hagemann,j.hertrich,steidl\}@math.tu-berlin.de.} \and Johannes Hertrich\samethanks \and Gabriele Steidl\samethanks}

  \theoremstyle{plain}
  \newtheorem{theorem}{Theorem}[section]
\newtheorem{lemma}[theorem]{Lemma}

  \newtheorem*{theorem*}{Theorem}
  \newtheorem*{lemma*}{Lemma}
  \newtheorem*{proposition*}{Proposition}
  \newtheorem*{corollary*}{Corollary}
  \newtheorem*{conjecture*}{Conjecture}

  \theoremstyle{definition}
  
  \newtheorem{example}[theorem]{Example}
  \newtheorem{remark}[theorem]{Remark}

\begin{document}

\maketitle

\begin{abstract}
Normalizing flows, diffusion normalizing flows and variational autoencoders are powerful generative models.
This chapter provides a unified framework to handle these approaches via Markov chains. 
We consider stochastic normalizing flows as a pair of Markov chains 
fulfilling some properties and show how many state-of-the-art models for data generation fit 
into this framework.
Indeed numerical simulations show that including stochastic layers improves the expressivity of the network and allows
for generating multimodal distributions from unimodal ones. 
The Markov chains point of view enables us 
to couple both deterministic layers as invertible neural networks
and stochastic layers 
as Metropolis-Hasting layers, Langevin layers, variational autoencoders and diffusion normalizing flows in a mathematically sound way. 
Our framework establishes a useful mathematical tool to combine the various approaches.
\end{abstract}

\section{Introduction}
Generative models have seen tremendous success in the recent years as they are able to produce diverse and very high-dimensional images. In particular Variational autoencoders (VAEs) were among the first ones to produce high quality samples from complex image distributions.
VAEs were originally introduced in \cite{KW2013}
and have seen a large number of modifications and improvements for quite different applications.
For some overview on VAEs, we refer to \cite{Kingma_2019}.
Recently, diffusion normalizing flows arising from the Euler discretization of a certain 
stochastic differential equation were proposed in \cite{zhang2021diffusion}.
On the other hand, normalizing flows 
including invertible residual neural networks (ResNets) considered in \cite{BGCDJ2019,CBDJ2019,HZRS2016},
invertible neural networks, see, e.g., \cite{AKRK2019,DSB2017,nice2014,kingma2018glow,MMRGN2018, RM2015} 
and autoregessive flows examined in \cite{CTA2019,DBMP2019,huang2018neural,PPM2017} 
are a popular classes of generative models. In contrast to VAEs, normalizing flows are explicitly invertible, which allows for exact likelihood computations.

In this tutorial, we will use finite normalizing flows which are
basically concatenations of diffeomorphisms with tractable Jacobian determinant. However, a quick explanation on continuous normalizing flows \cite{CRBD2018} will be given in the appendix. 
For a general overview of the deep generative modelling landscape, we refer to \cite{RH2021} and the references therein. 

\begin{figure}
\begin{center}
\includegraphics[width=\textwidth]{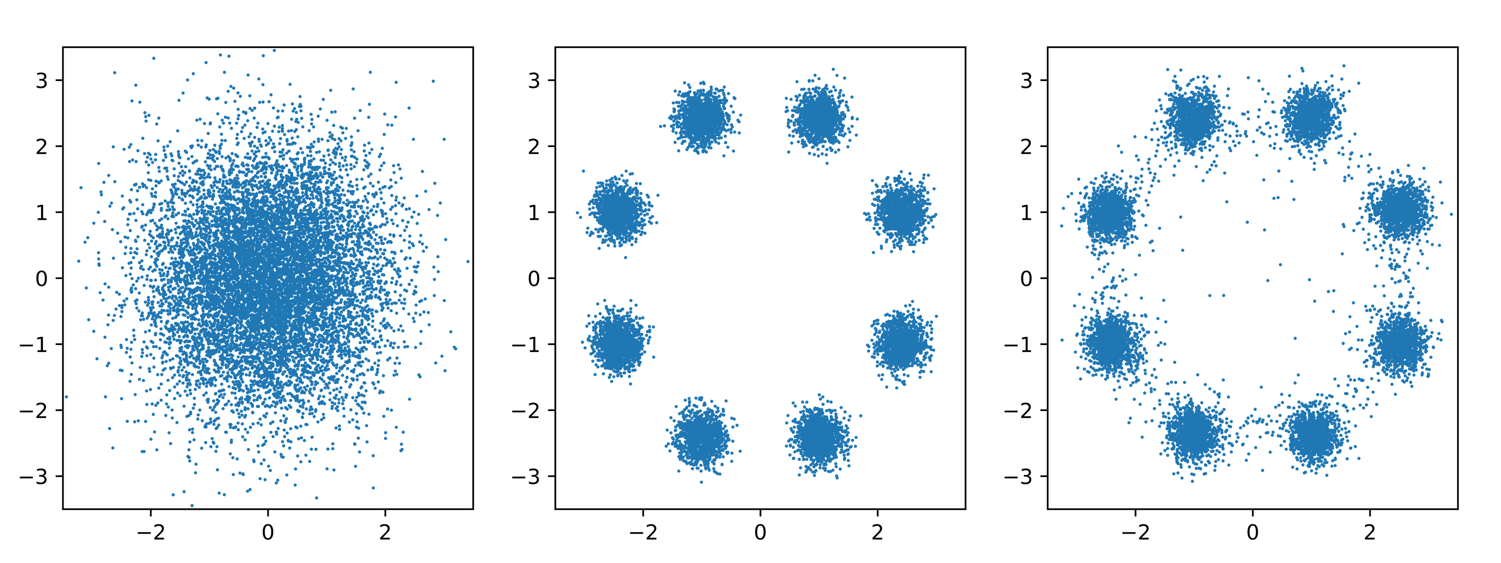}
\end{center}
\caption{Lack of expressiveness of invertible neural network.}
\label{fig:gaussian}
\addnote{
Left: Gaussian standard distribution (latent distribution),
Middle: Gaussian mixture distribution with colors (target distribution),
Right: Results of an invertible neural network trying to generate the multimodal distribution
from the unimodal Gaussian one.}
\end{figure}

Unfortunately, invertible neural networks suffer from a limited expressiveness.
More precisely, their major drawbacks  are topological constraints, 
see, e.g. \cite{Falorsietal2018,FHDF2019}, which means that the topological shape of latent and target distributions should "match" \cite{CCDD19}. In fact, if the pushforward of a unimodal distribution under a continuous map remains connected so it cannot represent a "truly" multimodal distribution perfectly. 
It was shown in \cite{HN2021}, see also \cite{behrmann2020understanding,CCDD19}, 
that for an accurate match between such distributions, the Lipschitz constant 
of the inverse flow has to approach infinity. Similar difficulties appear when mapping
to heavy-tailed distributions as observed in \cite{JKYB20}. 
See Figure \ref{fig:gaussian} for a typical example.

\begin{figure}
\begin{center}
\includegraphics[width=.3\textwidth]{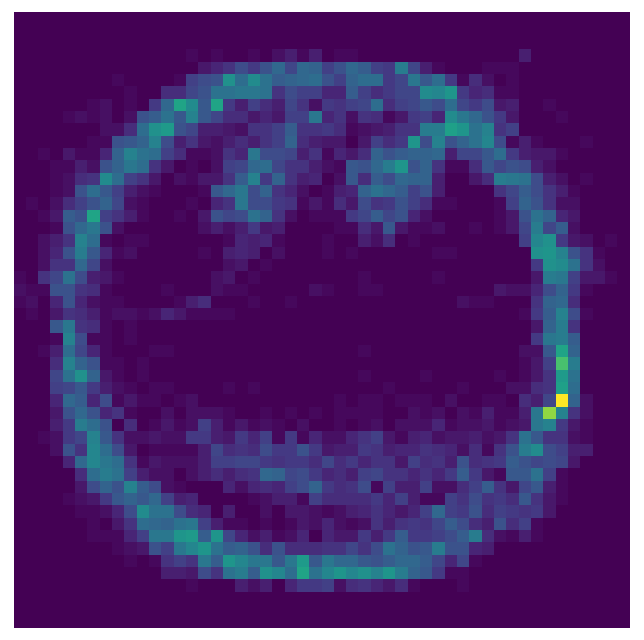}
\hspace{1cm}
\includegraphics[width=.3\textwidth]{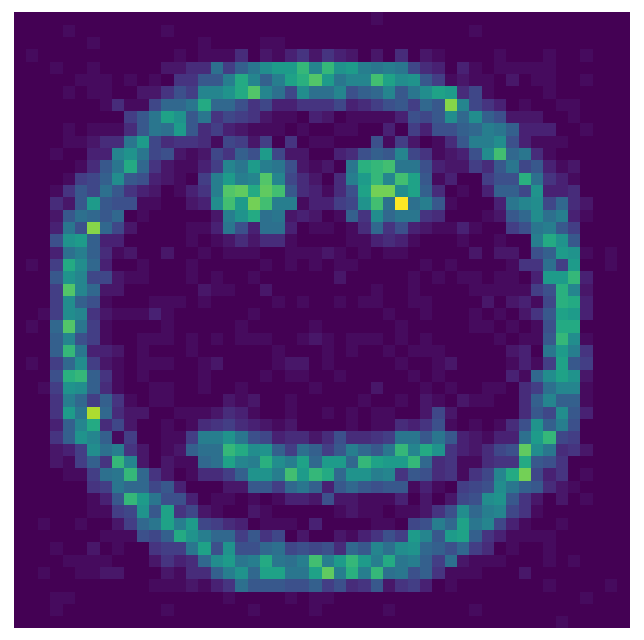}
\end{center}
\caption{Lack of expressiveness of normalizing flows as opposed to stochastic normalizing flows.}
\label{fig:smileys}
\addnote{A smiley modeled as 2d density generated by a normalizing flow on the left and with stochastic steps on the right. }
\end{figure}

To improve the expressiveness of normalizing flow architectures,
the authors of \cite{WKN2020} introduced stochastic normalizing flows. These are a generalization of deterministic flow layers and stochastic sampling methods, which allows for a use of both within one framework. 
One way to think about these methods is that they force the flow to obey a certain path: Often, for MCMC or Langevin methods, 
one needs to specify a density to anneal to, so that we define a path on which the flow moves to the target density, which follows ideas in \cite{WKN2020,Neal2001}.
The advantages can be seen in Figure \ref{fig:smileys}. Here on the left there is the modeled density 
of an image as a 2d density with stochastic normalizing flow versus a standard invertible neural network on the right.
The stochastic steps enable to model high concentration regions much better without smearing leaving connections between modes.
However, it turns out that also variational autoencoder can be modeled as stochastic layers. 
For those layers, we do not need to define interpolating densities, we only need to relax our notion of invertibility to "probabilistic" invertibility. 

In \cite{HHS2021} we considered 
stochastic normalizing flows from a Markov chain point of view.
In particular, we replaced the transition densities 
by general Markov kernels and provided mathematically sound derivations using
Radon-Nikodym derivatives. This allowed to incorporate deterministic flows as well as Metropolis-Hasting flows
which do not have densities into the mathematical framework.

The aim of this tutorial is to propose the sound framework of Markov chains to combine \emph{deterministic} normalizing
and \emph{stochastic} flows, in particular VAEs, diffusion normalizing flows and MCMC layers. 
It is addressed to readers with a basic background in measure theory
who are interested in generative models in machine learning and who want to see the clear mathematical relations between various methods appearing in the literature.

More precisely, we establish a pair of Markov chains that are inverse to each other in a broad sense we will explain later. 
This provides a powerful tool for coupling different architectures, which are used in many places in the machine learning literature. However, viewing them through the lens of stochastic normalizing flows a lot of very recent ideas can be \emph{unified} and \emph{subsumed} under this notion, which is quite elegant and useful. This gives a tool to combine variational autoencoder, diffusion layers which are inspired by stochastic differential equations, coupling based invertible neural network layer and stochastic MCMC methods all in one. Furthermore, we will provide a loss function that is an upper bound to the Kullback--Leibler distance between target and sampled distribution, 
enabling \emph{simultaneous} training of all those layers in a data driven fashion. 
We will demonstrate the universality of this framework by applying this to three inverse problems. The inverse problems consist of conditional image generation via 2d densities, a high-dimensional mixture problem with analytical ground truth and a real world problem coming from physics. 
The code for the numerical examples is available online\footnote{\url{https://github.com/PaulLyonel/Gen_norm_flow}}.

\subsection*{Related work}
Among the first authors who introduced stochastic diffusion like steps for forward 
and backward training of Markov kernels were the authors of \cite{SWMG2015}. 
Further, stochastic normalizing flows are closely related to the 
so-called nonequilibrium candidate Monte Carlo  method from 
nonequilibrium statistical mechanics introduced by \cite{NCMC2011}, in which deterministic layers are combined with 
stochastic acceptance-rejection steps with the difference that the deterministic steps are given beforehand by the 
physical example.
Furthermore, the authors of \cite{arbel2021annealed} also use MCMC and importance layers between normalizing flow layers, but as a 
difference to stochastic normalizing flows each of the flow layers is optimized via a layerwise loss with the backward 
Kullback--Leibler divergence. This avoids some of the gradient issues of stochastic normalizing flows.

Relations between normalizing flows and other approaches as VAEs 
were already mentioned in the literature.
So there exist several works 
which model the latent distribution of a VAE by 
normalizing flows, see \cite{DW2019,RM2015},
or 
by stochastic differential equations see \cite{VKK2021}.
Using the Markov chain derivation, all of these models can be share a lot of similarities with stochastic normalizing flows, 
even though some of them employ different training techniques for minimizing the loss function.
Further, the authors of \cite{GSS2019} modified the learning of the covariance matrices 
of decoder and encoder of a VAE using normalizing flows. 
This can also be viewed as one-layer stochastic normalizing flow.
A similar idea was applied by \cite{LW17}, where the weight distribution of a 
Bayesian neural network by a normalizing flow was modeled.
To bridge the gap between VAEs and flows, the authors of \cite{DJHWW2020} introduce injective and surjective layers and call them Sur{VAE}s. They introduce a variety of layers to make use of special structure of the data, such as permutation invariance as well as categorical values. This paper makes use of similar ideas as stochastic normalizing flows and also implemented variational autoencoder layer within the normalizing flows framework.

Further, to overcome the problem of expensive training in high dimensions, some recent papers as, e.g., 
\cite{cunningham2020normalizing,kothari2021trumpets} 
propose also other combinations of a dimensionality reduction and normalizing flows. 
The construction from \cite{cunningham2020normalizing} can be viewed as a variational autoencoder 
with special structured generator and can therefore be considered as one-layer stochastic normalizing flow. 
For a recent application of VAEs as priors in inverse problems in imaging we refer to \cite{GAT2021}.
Finally, the authors of \cite{kothari2021trumpets} proposed 
to reduce the dimension in a first step by a non-variational autoencoder 
and the optimization of a normalizing flow in the reduced dimensions in a second step.
\newpage

\section{Preliminaries} \label{sec:prelim}

\paragraph{Basics of probability}
Let $(\Omega,\mathcal A, \mathbb P)$ be a probability space.
By a  probability measure on $\R^d$ we always mean a 
probability measure defined on the Borel $\sigma$-algebra $\mathcal B(\R^d)$.
Let ${\mathcal P}(\mathbb R^d)$ denote the set of probability measures on  $\mathbb R^d$.
Given a random variable $X: \Omega \rightarrow \R^d$, 
we use the push-forward notation  (or image measure)
$$P_X = X_{\#} \mathbb P \coloneqq \mathbb P \circ X^{-1}$$ 
for the corresponding  measure on $\R^d$. 

\begin{figure}
\begin{center}
\begin{tabular}{c}
\includegraphics[width=.85\textwidth]{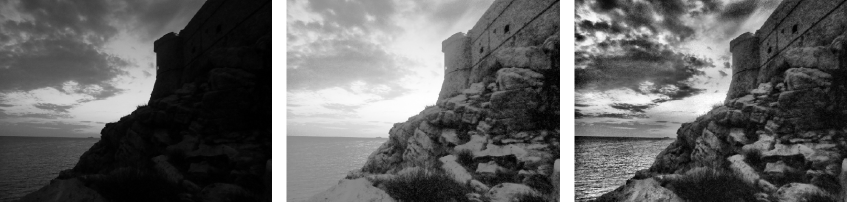}\\
\includegraphics[width=.85\textwidth]{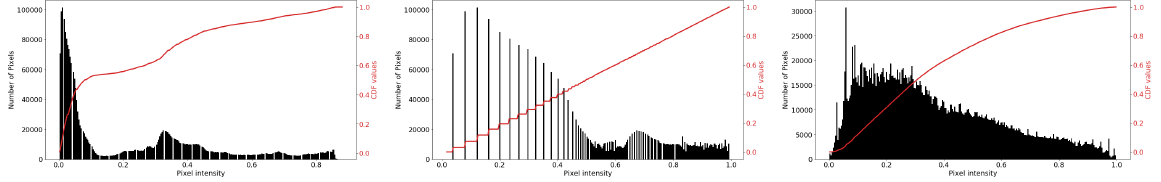}\\
Histograms of images as empirical densities and cumulative density function.\\
\end{tabular}
\vspace{0.5cm}

\begin{tabular}{c}
\includegraphics[width=.4\textwidth]{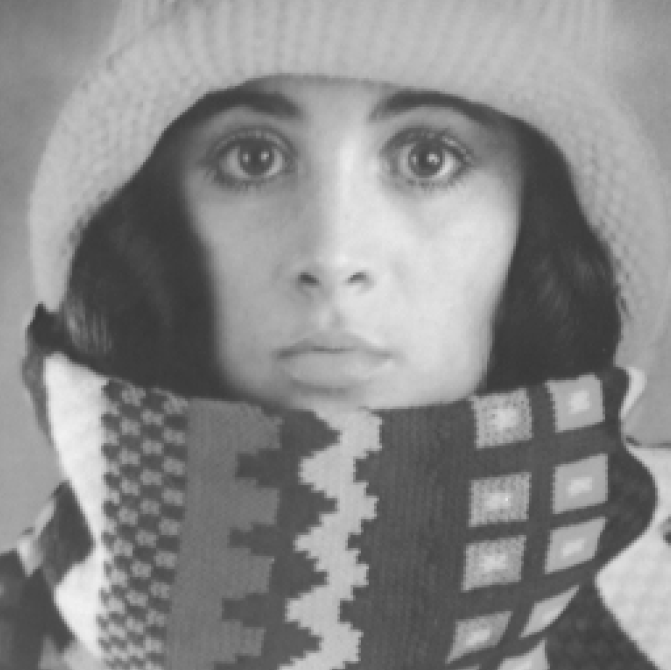} \hspace{0.5cm}
\includegraphics[width=.4\textwidth]{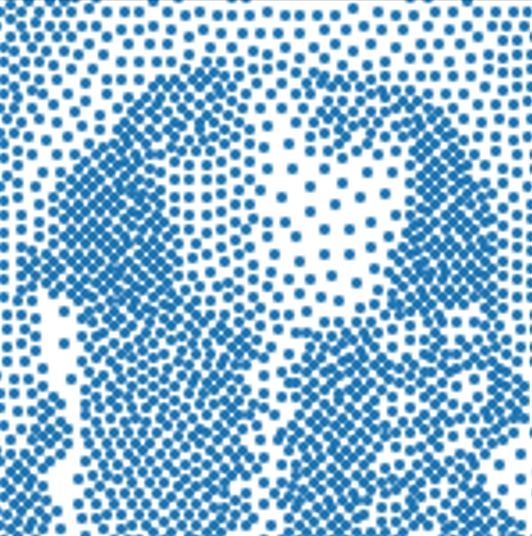}\\
Image as density function  on $\mathbb R^2$ and as discrete measure at stippled points.
\end{tabular}

\begin{tabular}{c}
\includegraphics[width=.8\textwidth]{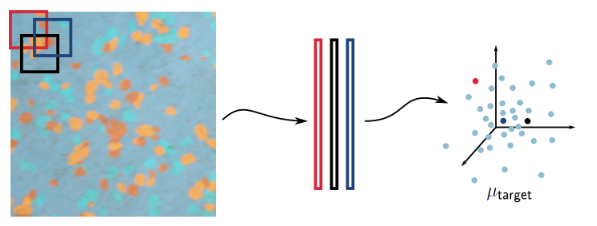}\\
Images patches of size $4 \times 4$ rearranged as empirical measure on $\mathbb R^{16}$ 
\end{tabular}
\end{center}
\caption{Different possibilities to assign measures to images as used in \cite{TeuSteGwoSchWei11}, \cite{HHR2021}, \cite{Hou21Patch}.} \label{fig:images}
\end{figure}

A measure $\mu \in \mathcal P(\R^d)$ is 
\emph{absolutely continuous} with respect to $\nu$ and we write $\mu \ll \nu$ 
if for every $B \in \mathcal B(\R^d)$ with $\nu(B) = 0$ we have $\mu(B) = 0$.
If $\mu, \nu \in \mathcal P(\R^d)$ satisfy $\mu \ll \nu$, then the \emph{Radon-Nikodym derivative} 
$\tfrac{\dx \mu}{\dx \nu} = \rho \in L^1(\R^d,\nu)$ exists 
and $\mu = \rho \nu$.
Special probability measures are 
\begin{itemize}
\item finite discrete measures
$$
\mu = \sum_{j=1}^N \mu_j \delta_{x_j}, \quad \mu_j >0 , \, \sum_{j=1}^N \mu_j = 1, \, x_j \in \mathbb R^d,
$$
where $\delta_{x}  (B) \coloneqq 1$ if $x \in B$ and $\delta_{x}  (B) \coloneqq 0$ otherwise.
If $\mu_j = \frac 1N$ for all $j=1,\ldots,N$, then these measures are also known as \emph{empirical measures} and  otherwise as
\emph{atomic measures}.
\item measures with densities, 
i.e. they are absolutely continuous with respect to the Lebesgue measure $\lambda$ and their density
is \smash{$\tfrac{\dx \mu}{\dx \lambda}$}.
\end{itemize}
Examples, showing why probability measures are interesting in image processing
are given in Figure \ref{fig:images}. It highlights different possibilities to
assign probability measures to images.

\paragraph{Markov Kernels} \label{sec:mc}
In this paragraph, we introduce a way to describe a generalized, random push forward via Markov kernels
see, e.g., \cite{LeGall, HHS2021}. This will be needed in order to describe the distribution, which is obtained by performing Markov Chain Monte Carlo layers.

A \emph{Markov kernel} 
$\mathcal K\colon \R^n\times \mathcal B(\R^d)\to [0,1]$ is a mapping such that
\begin{itemize}
	\item[i)]  $\mathcal K(\cdot,B)$ is measurable for any $B\in\mathcal B(\R^d)$, and 
	\item[ii)] $\mathcal K(x,\cdot)$ is a probability measure for any $x\in\R^n$.
\end{itemize}

For $\mu \in \mathcal P(\R^n)$, the measure $\mu\times \mathcal K$ on 
$\R^n\times\R^d$ is defined by
\begin{equation} \label{def}
	(\mu\times \mathcal K)(A\times B)\coloneqq \int_A \mathcal K(x,B) \dx \mu(x).
\end{equation}

Note that this definition captures all sets in $\mathcal B(\R^n \times \mathbb R^d)$ since
the measurable rectangles form a $\cap$-stable generator of $\mathcal B(\R^n\times\R^d)$.
Then, it holds for all integrable $f$ that
$$
\int_{\R^n\times\R^d}f(x,y) \dx (\mu\times \mathcal K)(x,y)=\int_{\R^n}\int_{\R^d}f(x,y) \dx \mathcal K(x,\cdot)(y) \dx \mu(x).
$$
Analogously to \eqref{def}, we define the product of a measure $\mu\in\mathcal P(\R^{d_0})$ and Markov kernels $\mathcal K_t\colon \R^{d_{t-1}}\times\mathcal B(\R^{d_t})\to[0,1]$ by
\begin{align*}
	&\quad(\mu\times\mathcal K_1\times\cdots\times\mathcal K_T)(A_0\times \cdots\times A_T)\\
	&\coloneqq \int_{A_0}\int_{A_1}\cdots\int_{A_{T-1}} \mathcal K(x_{T-1},A_T)\dx \mathcal K(x_{T-2},\cdot)(x_{T-1})\cdots\dx 
	\mathcal K(x_{0},\cdot)(x_{1})\dx\mu(x_0).
\end{align*}
As we will see later, this will be the notion to be used to describe the joint distributions of the measure $\mu$ and the measures obtained by iteratively applying the respective Markov kernels. This will be crucial for the definition and minimization of stochastic normalizing flows.

In the following, we use the notion 
of the \emph{regular conditional distribution} of a random variable $X$ 
given a random variable $Y$ which is defined as the $P_Y$-almost 
surely unique Markov kernel $P_{Y|X=\cdot}(\cdot)$ with the property
\begin{equation} \label{sense}
	P_{X}\times P_{Y|X=\cdot}(\cdot)=P_{(X,Y)}.
\end{equation}
We will use the abbreviation $P_{Y|X} = P_{Y|X=\cdot}(\cdot)$ if the meaning is clear from the context.

A sequence $(X_0,\ldots,X_T)$, $T \in \mathbb N$ of $d_t$-dimensional random variables $X_t$, $t=0,...,T$, is called a \emph{Markov chain},
if there exist Markov kernels  
$$\mathcal K_t = P_{X_t|X_{t-1}}\colon \R^{d_{t-1}}\times \mathcal B(\R^{d_t})\to [0,1]$$
in the sense \eqref{sense}
such that it holds
\begin{align}\label{eq_path_measure}
	P_{(X_0,...,X_T)} = P_{X_0} \times P_{X_1|X_{0}}  \times \cdots \times P_{X_T|X_{T-1}}.
\end{align}
The Markov kernels $\mathcal K_t$ also called \emph{transition kernels}.
If the measure $\mathcal K_t(x,\cdot) = P_{X_t|X_{t-1}=x}$ has a density $k_t(x,y)$, 
and $P_{X_{t-1}}$ resp. $P_{X_t}$ have densities $p_{{X_{t-1}}}$ resp. $p_{X_t}$, then setting $A \coloneqq \mathbb R^{d_{t-1}}$
in equation \eqref{def} results in
\begin{equation} \label{def_dichte}
	p_{X_t}(y) = \int_{\mathbb R^{d_{t-1}}} k_t(x,y) p_{{X_{t-1}}}(x) \dx x.
\end{equation}

In this tutorial we will use two ,,distance'' functions on the space of probability measures,
namely the Kullback-Leibler divergence and the Wasserstein-1 distance.

\paragraph{Kullback--Leibler divergence}
For $\mu,\nu\in {\mathcal P}(\R^d)$ with existing Radon-Nikodym derivative 
$\rho = \frac{\dx \mu}{\dx \nu}$ of $\mu$ with respect to $\nu$, 
the \emph{Kullback-Leibler divergence} is defined by
\begin{equation} \label{KLdef}
\mathrm{KL} (\mu,\nu) \coloneqq \int_{\mathbb R^d} \log \rho  \, \dx \mu(x).
\end{equation}
In case that the above Radon-Nikodym derivative does not exist, we set $\mathrm{KL} (\mu,\nu) \coloneqq + \infty$.
The Kullback-Leibler divergence is neither symmetric nor fulfills a triangular inequality, 
but it holds 
$\mathrm{KL} (\mu,\nu) = 0$ 
if and only if $\mu = \nu$. 
In particular, we have for measures $\mu,\nu \in \mathcal{P}(\R^d)$ which are absolutely continuous 
with respect to the Lebesgue measure
with densities $\rho_\mu, \rho_\nu$ that
$$
\mathrm{KL}(\mu,\nu) = \int_{\R^d} \log \left( \frac{\rho_\mu}{\rho_\nu} \right) \rho_\mu  \dx x,
$$
for discrete probability measures
$\mu = \sum_{j=1}^n \mu_j \delta_{x_j}$ and $\nu = \sum_{j=1}^n \nu_j \delta_{x_j}$ 
that
$$
\mathrm{KL}(\mu,\nu) = \sum_{j=1}^n \log \left( \frac{\mu_j}{\nu_j}\right) \mu_j.
$$
The Kullback-Leibler divergence does not depend on the geometry of the underlying space.
In contrast, the Wasserstein distance considered in the next paragraph takes spatial distances into account.

\paragraph{Wasserstein distance}
In the following, we revist Wasserstein distances, see e.g.~\cite{PC2019,Villani2003} for a more detailed overview.
For $p \in [1,\infty)$,
the \emph{$p$-Wasserstein distance}  $W_p$ between measures $\mu,\nu \in \mathcal P(\R^d)$ with finite $p$-th moments
is defined by
\begin{align} \label{eq:OTprimal}
W_p(\mu,\nu) \coloneqq \biggl( \inf_{\pi \in \Pi(\mu,\nu)} 
\int_{\R^d \times \R^d} \|x - y\|^p \mathrm{d} \pi(x,y) \biggr)^\frac{1}{p},
\end{align}
where $\Pi(\mu,\nu)$ denotes the measures on $\R^d \times \R^d$ with marginals $\mu$ and $\nu$.
It is a metric on the set of measures from $\mathcal P (\R^d)$ with finite $p$-th moments, 
which metrizes the weak topology, i.e., $\lim_{n \rightarrow \infty} W_p (\mu_n, \mu) = 0$
if and only if
$\mu_n \weakly \mu$
and 
$$
\int_{\mathbb R^d} \|x\|^p \dx \mu_n (x) \rightarrow \int_{\mathbb R^d} \|x\|^p \dx \mu (x)
$$
as $n \rightarrow \infty$.
For $1 \le p \le q < \infty$ it holds $W_p \le W_q$.
The distance $W_1$ is also called
\emph{Kantorovich-Rubinstein distance} or \emph{Earth's mover distance}.
Switching to the dual problem the Wasserstein-1 distance can be rewritten as
\begin{equation} \label{wasser_1_dual}
W_1(\mu,\nu)
= \max_{|\varphi|_{\Lip} \le 1}  \int_{\mathbb R^d} \varphi \, \dx (\mu -  \nu),
\end{equation}
where the maximum is taken over all Lipschitz continuous functions with Lipschitz constant bounded by 1.

\newpage
\section{Normalizing Flows} \label{sec:nf}
\begin{figure}
\centering
  \includegraphics[width=\textwidth]{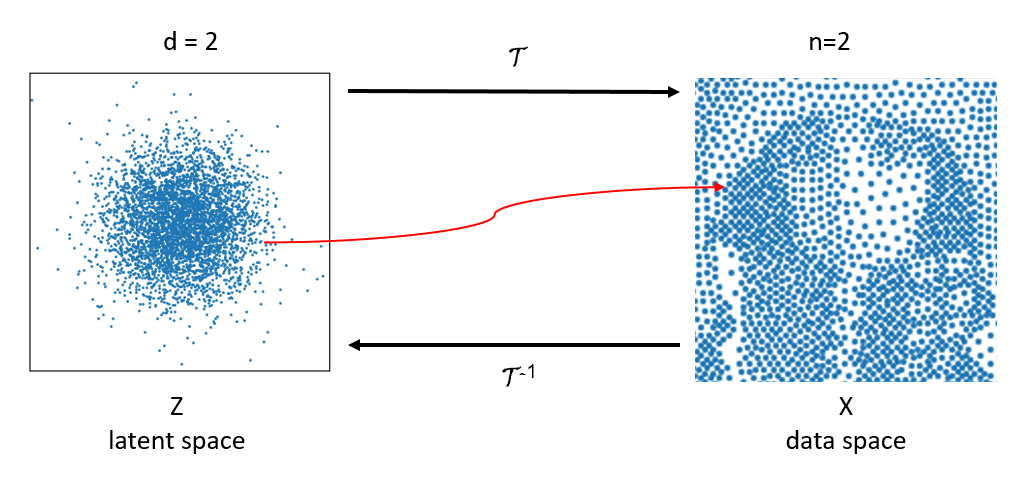}  
	\caption{\label{fig:1}
	Illustration of generative 	modeling.	}
	\addnote{The invertible NN $\mathcal T: \mathbb R^2 \rightarrow \mathbb R^2$ 
	maps samples from the latent standard normal distribution to the
	target distribution.}
\end{figure}

In this section, 
we give a quick overview how to train normalizing flows and how to interpret them as finite time Markov chains.
Normalizing flows are used to model 
a data distribution $P_X$ by 
the push-forward of a simpler latent distribution $P_Z$
by a diffeomorphism 
$\mathcal T_\theta\colon\R^d\to\R^d$, see \cite{pmlr-v37-rezende15}. 
Usually $P_Z$ is the standard normal distribution.
Each sample from the latent space gets mapped to a corresponding sample in data space.
For an illustration, see Figure \ref{fig:1}.

In this tutorial, $\mathcal T_\theta$ will follow a similar architecture as in \cite{DSB2017,AKRK2019}.
A brief description of such networks is given in the Appendix \ref{app:INN}.
For better readability, we skip the dependence of $\mathcal T_\theta$ on the parameter $\theta$ and write $\mathcal T = \mathcal T_\theta$.
We wish to learn $\mathcal T$ such that it holds 
$$
P_X \approx\mathcal T_\#P_Z,\quad\text{or equivalently}\quad P_Z\approx\mathcal T^{-1}_\#P_X.
$$
If $P_Z$ has a density $p_Z$, the following \emph{change of variables formula} for densities holds true:
\begin{equation} \label{push-forward}
	p_{T_\#P_Z} (x) = p_{Z} \big( \mathcal T^{-1} (x) \big) |\mathrm{det} \nabla \mathcal T^{-1} (x) |.
\end{equation}

The approximation can be done by minimizing the Kullback-Leibler divergence
\begin{align*}
\mathrm{KL}(P_X,\mathcal T_\#P_Z)
&=
\E_{x\sim P_X}\Big[\log\Big( \frac{ p_X}{ p_{\mathcal T_\#P_Z}} \Big)\Big]\\
&= 
\E_{x\sim P_X}\left[\log p_X \right]-\E_{x\sim P_X} \left[\log p_{\mathcal T_\#P_Z} \right]
\\
&=
\E_{x\sim P_X}\left[\log p_X \right]-
\E_{x\sim P_X} \left[\log  p_Z \circ \mathcal T^{-1}  \right] \\
&\quad - 
\E_{x\sim P_X} \left[\log |\mathrm{det}(\nabla \mathcal T^{-1} )| \right].
\end{align*}
Leaving out the constants not depending on $\theta$, we obtain the loss function
\begin{align} \label{loss_n}
\mathcal L_{\text{NF}} (\theta) &
\coloneqq -
\E_{x\sim P_X} \left[\log  p_Z \circ \mathcal T^{-1}  \right] - 
\E_{x\sim P_X} \left[\log |\mathrm{det}(\nabla \mathcal T^{-1} )| \right]\\
&= 
-\E_{x\sim P_X} \left[\log p_{\mathcal T_\#P_Z} \right]. \label{soo}
\end{align}
Note that this loss can now be optimized when samples from the distribution $P_X$ are available, as the $\log p_Z$ term is easy to evaluate for a standard normal density $p_Z$, and the log determinant of the flow is differentiable by construction. As can be seen in Appendix \ref{app:INN}, this is quite easy to evaluate for our choice of architecture.

\begin{remark}\label{rem_backward_KL}
As we already noted, the Kullback-Leibler divergence is not symmetric and hence one could in principle also consider $\mathrm{KL}(\mathcal T_\#P_Z,P_X)$. This would in fact yield the following loss (with same minimizer $\mathcal T$):
\begin{align*}
	\mathrm{KL}(\mathcal T_\#P_Z,P_X)
	&=
	\mathrm{KL}(P_Z,\mathcal{T}^{-1}_\#P_X)\\
	&= 
	\E_{z\sim P_Z}\left[\log p_Z \right]-\E_{z\sim P_Z} \left[\log p_{\mathcal{T}^{-1}_\#P_X} \right]
	\\
	&=
	\E_{z\sim P_Z}\left[\log p_Z \right]-
	\E_{z\sim P_Z} \left[\log  p_X \circ \mathcal T  \right] \\
	&\quad - 
	\E_{z\sim P_Z} \left[\log |\mathrm{det}(\nabla \mathcal T(z) )| \right].
\end{align*}
This loss function is usually called reverse or backward KL and has vastly different optimization properties compared to the forward  $\mathrm{KL}(P_X,\mathcal T_\#P_Z)$. Furthermore, it does not require samples from $P_X$, but instead it requires the evaluation of the energy or negative log of $p_X$, see \cite{KDSK2020}, which is not possible in many applications.
\end{remark}

The network $\mathcal T$ is constructed by concatenating smaller blocks 
$$\mathcal T=\mathcal T_T\circ\cdots\circ \mathcal T_1$$
which are invertible networks on their own.
Then, the blocks $\mathcal T_t\colon\R^d\to\R^d$ 
generate a pair of Markov chains $\big((X_0,...,X_T),(Y_T,...,Y_0)\big)$ by 
\begin{align*}
X_0\sim P_Z,\quad &X_t=\mathcal T_t(X_{t-1})\quad\text{and}\\
Y_T\sim P_X, \quad &Y_{t-1}=\mathcal T_t^{-1}(Y_t).
\end{align*}
Here, for all $t=0,...,T$, the dimension $d_t$ of the random variables $X_t$ and $Y_t$ is equal to $d$.

\begin{lemma} \label{kernel_MH}
The Markov kernels 
$\mathcal K_t = P_{X_t|X_{t-1}} \colon \R^{d}\times \mathcal B(\R^{d})\to [0,1]$ 
and
$\mathcal R_t = P_{Y_{t-1}|Y_t} \colon \R^{d}\times \mathcal B(\R^{d})\to [0,1]$
belonging to the above Markov chains are given by the Dirac distributions
\begin{equation} \label{kern_det}
\mathcal K_t(x,\cdot)=\delta_{\mathcal T_t(x)},\qquad
\mathcal R_t(x,\cdot)=\delta_{\mathcal T_t^{-1}(x)}. 
\end{equation}
\end{lemma}

\begin{proof}
For any $A,B \in \mathcal B(\mathbb R^d)$ it holds 
\begin{align} 
P_{(X_{t-1},X_{t})}(A\times B)
&=\int_{\mathbb R^d} 1_{A\times B}(x_{t-1},x_t) \dx P_{(X_{t-1},X_t)}(x_{t-1},x_t)\\
&=\int_{\mathbb R^d}  1_{A}(x_{t-1})1_B(x_t) \dx P_{(X_{t-1},X_t)}(x_{t-1},x_t).
\end{align}
Since $P_{(X_{t-1},X_t)}$ is by definition 
concentrated on the set $\{ \left( yx_{t-1},\mathcal T_t(x_{t-1}) \right):x_{t-1}\in\R^d\}$, this becomes
\begin{align}
P_{(X_{t-1},X_{t})}(A\times B)
&=\int_{\mathbb R^d \times \mathbb R^d}1_{A}(x_{t-1})1_B(\mathcal T_t(x_{t-1})) \dx P_{(X_{t-1},X_t)}(x_{t-1},x_t)\\
&=\int_A 1_B(\mathcal T_t(x_{t-1})) \dx P_{X_{t-1}}(x_{t-1})\\
&=\int_A \delta_{\mathcal T_t(x_{t-1})}(B) \dx P_{X_{t-1}}.
\end{align}
Consequently, by \eqref{def},
the transition kernel $\mathcal K_t=P_{X_t|X_{t-1}}$ is given by $\mathcal K_t(x,\cdot)=\delta_{\mathcal T_t(x)}$. 
\end{proof}

Due to their correspondence to the layers $\mathcal T_t$ and $\mathcal T_t^{-1}$ from the normalizing flow $\mathcal T$, 
we call the Markov kernels $\mathcal K_t$ \emph{forward layers}, while the Markov kernels $\mathcal R_t$ 
are called \emph{reverse layers}.

\newpage
\section{Stochastic Normalizing Flows}\label{sec:snf}
\begin{figure}
\begin{center}
\includegraphics[width=.95\textwidth]{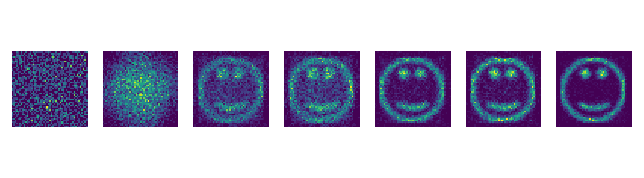}
\end{center}
\caption{The path on which the stochastic normalizing flow moves from the latent density (left most picture) to the target density (right most picture). We alternate invertible neural network layers with MCMC layers, starting with the first one.}
\label{path_snf}
\end{figure}

We have already seen that normalizing flows have limited expressiveness. 
The idea of stochastic normalizing flows is to replace some of the deterministic layers $\mathcal T_t$ from a normalizing flow 
by random transforms. 
From the Markov chains viewpoint, we replace the kernels $\mathcal K_t$ and $\mathcal R_t$ with the Dirac measure by more general Markov kernels.
In Figure \ref{path_snf} the interaction of stochastic steps in conjunction with deterministic ones is illustrated.
In particular, the stochastic steps effectively remove samples from low density regions.

Formally, a \emph{stochastic normalizing flow} (SNF) is a pair  $\big( (X_0,\ldots,X_T),(Y_T,...,Y_0) \big)$ of Markov chains of $d_t$-dimensional random variables $X_t$ and $Y_t$, $t=0,...,T$,
with the following properties:
\begin{itemize}
\item[] P1) $P_{X_t}, P_{Y_t}$ have the densities $p_{X_t}, p_{Y_t}\colon\R^{d_t}\to\R_{>0}$ for any $t=0,...,T$.
\item[] P2) There exist Markov kernels 
$\mathcal K_{t} = P_{X_t|X_{t-1}}$ and  
$\mathcal R_t = P_{Y_{t-1}|Y_t}$, $t=1,...,T$
such that 
\begin{align}
P_{(X_0,...,X_T)} &= P_{X_0} \times P_{X_1|X_{0}}  \times \cdots \times P_{X_T|X_{T-1}},\\ 
P_{(Y_T,...,Y_0)} &= P_{Y_T} \times P_{Y_{T-1}|Y_{T}} \times\cdots\times P_{Y_{0}|Y_{1}}.
\end{align}
\item[] P3) For $P_{X_t}$-almost every $x\in\R^{d_t}$, 
the measures $P_{Y_{t-1}|Y_t=x}$ and $P_{X_{t-1}|X_t=x}$ 
are absolutely continuous with respect to each other.
\end{itemize}
We say that the Markov chain $(Y_T,...,Y_0)$ is a \emph{reverse Markov chain} of $(X_0,\ldots,X_T)$,
see Lemma \ref{lem:reversal} in the appendix.
In applications, Markov chains usually start with a latent random variable 
$$X_0 = Z$$ 
on $\R^{d_0}$,
which is easy to sample from and we intend  to learn the Markov chain such that $X_T$ 
approximates a target random variable $X$ on $\R^{d_T}$, while
the reversed Markov chain is initialized with a random variable 
$$Y_T = X$$ 
from a data space and $Y_0$ should approximate the
latent variable $Z$.
As outlined in the previous paragraph, each deterministic normalizing flow is a special case of a SNF.

\subsection{Training SNFs}

We aim to find parameters of a SNF such that $P_{X_T}\approx P_X$. 
Recall, that for deterministic normalizing flows, it holds $P_{X_T}=\mathcal T_\#P_Z$, 
such that the loss function $\mathcal L_\text{NF}$ reads as $\mathcal L_\text{NF}=\mathrm{KL}(P_X,P_{X_T})$.
Unfortunately, the stochastic layers make it impossible to evaluate and minimize $\mathrm{KL}(P_X,P_{X_T})$.
Instead, we minimize the KL divergence of the joint distributions
$$
\mathcal L_\text{SNF}\coloneqq \mathrm{KL}(P_{(Y_0,...,Y_T)},P_{(X_0,...,X_T)}),
$$
which is an upper bound of $\mathrm{KL}(P_{Y_T},P_{X_T})=\mathrm{KL}(P_X,P_{X_T})$, 
see Lemma \ref{lem_KL_marginals} in the appendix.
The following theorem was proved in \cite{HHS2021}, Theorem 5.

\begin{theorem}
The loss function $\mathcal L_\text{SNF}$ can be rewritten as
\begin{align} 
\mathcal L_\text{SNF} (\theta) &= \mathrm{KL}(P_{(Y_0,...,Y_T)},P_{(X_0,...,X_T)})\label{eq_loss_SNF}
\\
&=
\E_{(x_0,...,x_T)\sim P_{(Y_0,...,Y_T)}}
\Big[\log\Big(\frac{p_X(x_T)}{p_{X_T}(x_T)}\prod_{t=1}^T f_t(x_{t-1},x_t)\Big)\Big]
\\
&=\E_{(x_0,...,x_T)\sim P_{(Y_0,...,Y_T)}}
\Big[\log\Big(\frac{p_X(x_T)}{p_Z(x_0)}\prod_{t=1}^T\frac{f_t(x_{t-1},x_t) p_{X_{t-1}(x_{t-1})}}{p_{X_{t}(x_{t})}}\Big)
\Big], 
\end{align}
where $f_t(\cdot,x_{t})$ is given by the Radon-Nikodym derivative 
$\frac{\dx P_{Y_{t-1}|Y_t=x_t}}{\dx P_{X_{t-1}|X_t=x_t}}$.
\end{theorem}

For the deterministic NF layers and the stochastic layers discussed in the next section
the quotients in the loss function \eqref{eq_loss_SNF}
are specified in Theorem~\ref{lem_logdets}. 
In particular, Theorem~\ref{lem_logdets} yields that we have for any deterministic normalizing flow 
that $\mathcal L_\text{NF}=\mathcal L_\text{SNF}$.

\section{Stochastic Layers}
In this section, we consider different stochastic layers, namely
\begin{itemize}
\item Langevin layer,
\item Metropolis-Hastings (MH) layer,
\item Metropolis-adjusted Langevin (MALA) layer,
\item VAE layer,
\item diffusion normalizing flow layer.
\end{itemize}
The first three layers were used, e.g. in \cite{HHS2021,WKN2020}. 
Further layers were introduced in \cite{DJHWW2020}.
 Note that the first three layers do not have trainable parameters.

In the following, let $\mathcal N(\mu,\Sigma)$ denote the normal distribution with density 
\begin{equation}\label{gaussian}
		\mathcal N (x;m,\Sigma) \coloneqq (2\pi)^{-\frac{d }{2}} |\Sigma|^{-\frac{1}{2}} 
		\,\exp\left(-\frac{1}{2}(x-m)^\tT \Sigma^{-1}(x-m) \right).
	\end{equation} 

\subsection{Langevin Layer}\label{sec_langevin_layer}
As for the deterministic layers we choose
$$d_{t-1}=d_t=d.$$ 
The basic idea is to push the distribution of $X_{t-1}$ into the direction 
of some proposal density $p_t\colon\R^{d}\to\R_{>0}$, 
whose choice is discussed in Remark~\ref{rem_prop_den} later.
We denote by 
$$u_t(x)\coloneqq-\log(p_t(x))$$ 
the negative log-likelihood of $p_t$.

To move in the direction of $p_t$, we follow the path of the so-called \emph{overdamped
Langevin dynamics}, i.e., the stochastic differential equation defined by
\begin{equation}\label{eq_lagevin_equation}
\dx L_s=-\nabla u_t(L_s)\dx t+\tfrac{2}{\beta} \dx B_s,
\end{equation}
with respect to the Brownian motion $B_s$ and damping constant $\beta>0$, see e.g.\ \cite{langevin_welling}.
It is known that this SDE admits the stationary distribution with unnormalized density $\exp(-\beta u_t)$ and that $L_s$ converges
in distribution to the distribution with density $p_t$, see e.g.~\cite{CK2012}.
In order to follow the path of \eqref{eq_lagevin_equation}, we use the explicit Euler discretization with step size $\eta$ given by
$$
L_{s+\eta}=L_s-\eta\nabla u_t(L_s)+\sqrt{\tfrac{2\eta}{\beta}} \xi_s,
$$
where $\xi_s\sim\mathcal N(0,I)$.

Using this motivation, we define the Langevin layer as the transition from $X_{t-1}$ to $X_t$ given by
\begin{equation}\label{eq_langevin_layer}
X_t \coloneqq X_{t-1}-a_1 \nabla u_t(X_{t-1})+a_2\xi_t,
\end{equation}
where $a_1,a_2>0$ are some predefined constants depending on the step size $\eta$ and the damping constant $\beta$
and $\xi_t\sim \mathcal N(0,I)$ such that $\sigma(\xi_t)$ 
and $\sigma \left( \cup_{s\le t-1} \sigma( X_s) \right)$ are independent.

Note that the Langevin layer \eqref{eq_langevin_layer} contains by definition a gradient ascent with respect to the  
log-likelihood of $p_t$. Therefore, it will remove samples from regions with a low density $p_t$.

Now, the corresponding Markov kernel $\mathcal K_t = P_{X_t|X_{t-1}}$ can be deduced by the following lemma.
As reverse layer, we use the same Markov kernel as the forward layer, i.e., 
$$\mathcal R_t=\mathcal K_t.$$

\begin{lemma} \label{kernel_langevin}
The Markov kernel $\mathcal K_t = P_{X_t|X_{t-1}} \colon \R^{d}\times \mathcal B(\R^{d})\to [0,1]$ 
belonging to the Langevin transition is 
\begin{align} \label{eq_langevin_kernel}
\mathcal K_t(x,\cdot) &\coloneqq \mathcal N(x-a_1\nabla u_t(x),a_2^2 I).
\end{align}
\end{lemma}

\begin{proof}
To determine the corresponding kernel, we use the
the independence of $\xi_t$ of $X_t$ and $X_{t-1}$ to obtain that $X_t$ and $X_{t-1}$ have the common density
\begin{align}
p_{(X_{t-1},X_t)}(x_{t-1},x_t)&=p_{X_{t-1},\xi_t}\big(x_{t-1},\tfrac1{a_2}(x_t-x_{t-1}+a_1\nabla u_t(x_{t-1})\big)\\
&=p_{X_{t-1}}(x_{t-1})p_{\xi_t}\big(\tfrac1{a_2}(x_t-x_{t-1}+a_1\nabla u_t(x_{t-1}))\big)\\
&=p_{X_{t-1}}(x_{t-1})\mathcal N(x_t;x_{t-1}-a_1\nabla u_t(x_{t-1}),a_2^2 I).
\end{align}
Then, for $A,B \in \mathcal B(\mathbb R^{d})$, it holds
\begin{align}
P_{(X_{t-1},X_t)}(A\times B)
&=\int_{A\times B} p_{X_{t-1}}(x_{t-1})\mathcal N(x_t;x_{t-1}-a_1\nabla u_t(x_{t-1}),a_2^2 I) \dx (x_{t-1},x_t)\\
&=\int_A \int_B\mathcal N(x_t;x_{t-1}-a_1\nabla u_t(x_{t-1}),a_2^2 I)\dx x_t p_{X_{t-1}}(x_{t-1})\dx x_{t-1}\\
&=\int_A  \mathcal K_t(x_{t-1},B) p_{X_t}(x_t)\dx P_{X_{t-1}}(x_{t-1}),
\end{align}
where 
\begin{align}
\mathcal K_t(x,\cdot) \coloneqq \mathcal N(x-a_1\nabla u_t(x),a_2^2 I).
\end{align}
By \eqref{def} and \eqref{sense} this is the Langevin transition kernel $P_{X_t|X_{t-1}}$.
\end{proof}
%
\subsection{Metropolis-Hastings Layer}\label{sec_MH_layer}
Again we choose
$d_{t-1}=d_t=d$
and 
$$\mathcal R_t=\mathcal K_t$$
and push the distribution of $X_{t-1}$ into the direction 
of some proposal density $p_t\colon\R^{d}\to\R_{>0}$.

The Metropolis-Hastings algorithm outlined in Alg. \ref{alg:MH} 
is a frequently used Markov Chain Monte Carlo type algorithm to sample
from a proposal distribution $P$ with known proposal density $p$, 
see, e.g.,\ \cite{robertsrosenthalMCMC}.
Under mild assumptions, the corresponding Markov chain $(X_k)_{k \in \mathbb N}$ admits the unique stationary distribution $P$ 
and $P_{X_k}\to P$ as $k \rightarrow \infty$ in the total variation norm, see, e.g. \cite{TLA2020}.

%
\begin{algorithm}
\begin{algorithmic}
		\State \textbf{Input:} $x_0 \in \mathbb R^d$, proposal density $p_t\colon\R^{d}\to\R_{>0}$
		\State \textbf{For} $k=0,1,\ldots$ do
		\State Draw $x'$ from $\mathcal N (x_k,\sigma^2 I)$ and $u$ uniformly in $[0,1]$.
    \State Compute the acceptance ratio
$$
\alpha(x_k,x') \coloneqq \min \left\{1, \frac{p(x')}{p(x_k)} \right\}.
$$
\State Set
$$
x_{k+1} \coloneqq \left\{
\begin{array}{ll}
x' & \mathrm{if} \;  u<\alpha(x_k,x'),\\
x_k & \mathrm{otherwise}.
\end{array}
\right.
$$
\State \textbf{Output:} $\{x_k\}_k$
\caption{Metropolis-Hastings Algorithm}
\label{alg:MH}
\end{algorithmic}
\end{algorithm}
%

In the MH layer, the transition from $X_{t-1}$ to $X_t$ is one step of a Metropolis-Hastings algorithm. 
More precisely,
let $\xi_t\sim\mathcal N(0,\sigma^2 I)$ and $U\sim\mathcal U_{[0,1]}$ be random variables such that
$\left( \boldsymbol{\sigma}(\xi_t),\boldsymbol{\sigma}(U),\boldsymbol{\sigma} \left( \bigcup_{s\le t-1} \boldsymbol{\sigma}( X_s) \right) \right)$
are independent.
Here $\boldsymbol{\sigma}(X)$ denotes the smallest $\boldsymbol{\sigma}$-al\-ge\-bra
generated by the random variable $X$.
Then,  we set
\begin{align}
X_t
&\coloneqq
X_{t-1} +1_{[U,1]} \left( \alpha_t( X_{t-1},X_{t-1}+\xi_t) \right) \, \xi_t
\end{align}
where
$$\alpha_t (x,y) \coloneqq \min \left\{1, \frac{p_t(y)}{p_t(x)} \right\}$$
with a proposal density $p_t$ which is discussed in Remark~\ref{rem_prop_den}.

Intuitively, the MH layer perturbs  a sample with some noise. Afterwards, we compare the proposal probabilites
 of the perturbed sample and the original sample. If the probability of the
perturbed sample is higher, we accept it with a high probability, other-
wise we reject it. Consequently, we remove samples from regions with
low proposal density.

The corresponding Markov kernel can be computed by the following lemma.

\begin{lemma} \label{MH_kernel_1}
The Markov kernel $\mathcal K_t = P_{X_t|X_{t-1}}\colon \R^{d}\times \mathcal B(\R^{d})\to [0,1]$ 
belonging to the Metropolis-Hastings transition is  
\begin{align} \label{kernel_MCMC}
\mathcal K_t(x,A)
&\coloneqq
\int_A \mathcal N(y;x,\sigma^2 I) \alpha_t (x,y) \dx y\\
& \quad +
\delta_x (A) \int_{\R^{d}} \mathcal N(y;x,\sigma^2 I) \left( 1- \alpha_t (x,y) \right) \dx y .
\end{align}
\end{lemma}

The proof is a special case of Lemma \ref{MH kernel} in the appendix.

\subsection{Metropolis-adjusted Langevin Layer}
Another kind of MH layer comes from the Metropolis-adjusted Langevin algorithm (MALA), 
see \cite{mala_layer,GC2011,RT1996}.
It combines the
Langevin layer from Section~\ref{sec_langevin_layer} with the Metropolis Hastings layer from Section~\ref{sec_MH_layer}.
Again we choose
$d_{t-1}=d_t=d$
and 
$$\mathcal R_t=\mathcal K_t$$
and push the distribution of $X_{t-1}$ into the direction 
of some proposal density $p_t\colon\R^{d}\to\R_{>0}$.
Let $\xi_t\sim\mathcal N(0,I)$,  $a_1,a_2>0$ and $u_t \coloneqq -\log p_t$ as in the Langevin
layer. Further, we choose
$$
    Q_t(x,\cdot)\coloneqq\mathcal N(x-a_1\nabla u_t(x),a_2^2 I),
		\qquad 
		q(\cdot|x)\coloneqq\mathcal N(\cdot|x-a_1\nabla u_t(x),a_2^2 I).
    $$
Then the Metropolis-adjusted Langevin algorithm is detailed in Alg. \ref{alg:MH_L}.

\begin{algorithm}
\begin{algorithmic}
		\State \textbf{Input:} $x_0 \in \mathbb R^d$, proposal density $p_t\colon\R^{d}\to\R_{>0}$, $a_1,a_2>0$  
		\State  \textbf{For} $k=0,1,\ldots$ do
		\State Draw $x'$ from $\mathcal N (x_k - a_1 \nabla u_t(x_k), a_2^2 I)$ and $u$ uniformly in $[0,1]$.
    \State Compute the acceptance ratio
$$
\alpha(x_k,x') \coloneqq \min \left\{1, \frac{p(x')q(x_k|x')}{p(x_n)q(x'|x_k)} \right\}.
$$
\State Set
$$
x_{k+1} \coloneqq \left\{
\begin{array}{ll}
x' & \mathrm{if} \;  u<\alpha(x_k,x'),\\
x_k & \mathrm{otherwise}.
\end{array}
\right.
$$
\State \textbf{Output:} $\{x_k\}_k$
\caption{Metropolis adjusted Langevin Algorithm}
\label{alg:MH_L}
\end{algorithmic}
\end{algorithm}

In the MALA layer, the transition from $X_{t-1}$ to $X_t$ is one step of a MALA algorithm. 
More precisely,  we set
\begin{align}
    X_t &\coloneqq X_{t-1} + 
		1_{[U,1]} \left( \alpha_t( X_{t-1},X_{t-1}-a_1\nabla u_t(X_{t-1}) +a_2\xi_t) \right)
				\left(a_2\xi_t-a_1\nabla u_t(X_{t-1}) \right),
    \end{align}
where
$\alpha_t$ is defined as in the MALA algorithm. 
As a combination of the Langevin layer and MH layer, the MALA layer also removes samples from regions with
small proposal density $p_t$.
The corresponding Markov kernel is determined by the following lemma.

\begin{lemma} \label{MALA_kernel_1}
The Markov kernel 
$\mathcal K_t = P_{X_t|X_{t-1}}
\colon \R^{d}\times \mathcal B(\R^{d}) \to [0,1]$ 
belonging to the MALA transition is 
\begin{align} \label{kernel_MALA}
\mathcal K_t(x,A)
&\coloneqq
\int_A q_t(y|x) \alpha_t (x,y) \dx y\\
& \quad +
\delta_x (A) \int_{\R^{d}} q_t(y|x) \left( 1- \alpha_t (x,y) \right) \dx y .
\end{align}
\end{lemma}

The proof is a special case of Lemma \ref{MH kernel} in the appendix.

\begin{remark}[Proposal Densities]\label{rem_prop_den}
The first three of these layers are based on stochastic sampling methods.
The idea of these layers is to push the distribution of $X_{t-1}$ into the direction
of some \emph{proposal density}, which has to be known up to a multiplicative constant a priori. Clearly, the interpolation of the proposal densities has its drawbacks \cite{grosse2013_intermediate}.

In the following, we address the question, how this proposal density can be chosen.
Recall that the random variable $X_0$ follows the latent density $P_Z$ and that $X_T$ approximates the target distribution $P_X$.
Moreover, for plenty of applications, the density $p_X$ is known up to a multiplicative constant, but classical sampling methods
as rejection sampling and MCMC methods take too much time and too many resources to be applicable. 
In this case, it appears to be reasonable to choose the proposal density $p_t$ as an interpolation between $p_Z$ and $p_X$.
Because of its simple computation in log-space, in literature mostly the geometric mean 
$p_t\propto p_Z^{(T-t)/T}p_X^{t/T}$ is used, see \cite{HHS2021,Neal2001,WKN2020}.

If the density $p_X$ is unknown, 
we can replace $p_X$ by some proper approximation.
This could be taken from probability densities 
$c\exp(-\lambda \mathcal R)$ which involve
regularizing terms $\mathcal R$ often used in variational image processing,
as e.g.~in \cite{AH2022}. For other
data driven regularizers see  \cite{ADHHMS2022,KEKP2020,LOS2018}.

In the case of Langevin layers, not the proposal density $p_t$ itself is required, but only the gradient of its logarithm.
Then, we can estimate the gradient of $\log(p_X)$ from the given samples of $X$ using score-matching, see e.g.\ \cite{SE2019}.
\hfill$\square$
\end{remark}

\subsection{VAE Layer} \label{sec:vae_as_snf}

In this section, we introduce variational autoencoders (VAEs) as another kind of stochastic layers 
of a SNF. First, we briefly revisit the definition of autoencoders and VAEs. 
Afterwards, we show that a VAE can be viewed as a one-layer SNF.

\paragraph{Autoencoders}
Autoencoders are a dimensionality reduction technique 
inspired by the principal component analysis. For an overview, see, e.g. \cite{GBC2016}.
For $d>n$, 
an autoencoder is a pair $(E,D)$ of neural networks, consisting
of an encoder $E=E_\phi\colon\R^d\to\R^n$ and a decoder $D=D_\theta\colon\R^n\to\R^d$, 
where $\theta$ and $\phi$ are the neural networks parameters. 
The network $E$ aims to encode samples $x$ from a $d$-dimensional distribution $P_X$ in the 
lower-dimensional space $\R^n$ such that the decoder $D$ is able to reconstruct them.
Consequently, it is a necessary assumption that the distribution $P_X$ is approximately concentrated on a
$n$-dimensional manifold.
A possible loss function to train $E_\phi$ and $D_\theta$ is given by
$$
\mathcal L_\text{AE}(\phi,\theta) \coloneqq \E_{x\sim P_X}[\|x-D_\theta(E_\phi(x))\|^2].
$$
Using this construction, autoencoders have shown to be very powerful for reduce the dimensionality of very complex datasets.

\paragraph{Variational Autoenconders via Markov Kernels}

Variational autoencoders (VAEs) orginally introduced by \cite{KW2013}, aim to use the power of autoencoders to approximate a probability distribution $P_X$ with density $p_X$ using a simpler distribution $P_Z$ with density $p_Z$ which is usually the standard normal distribution.
Here, the idea is to learn random transforms that push the distribution $P_X$ onto $P_Z$ and vice versa.
Formally, these transforms are defined by the Markov kernels
\begin{equation}
\mathcal K(z,\cdot)\coloneqq\mathcal N(\mu_\theta(z),\Sigma_\theta(z))
\quad\text{and}\quad
\mathcal R(x,\cdot)
\coloneqq
\mathcal N(\mu_\phi(x),\Sigma_\phi(x)),\label{eq_VAE_kernels}
\end{equation}
where 
$$D(z)=D_\theta(z)\coloneqq(\mu_\theta(z),\Sigma_\theta(z))$$ 
is a neural network with parameters $\theta$, 
which determines the parameters of the normal distribution within the definition of $\mathcal K$. 
Similarly, 
$$E(x)=E_\phi(x)\coloneqq(\mu_\phi(x),\Sigma_\phi(x))$$ 
determines the parameters within the definition of $\mathcal R$. 
In analogy to the autoencoders in the previous paragraph, $D$ and $E$ are called stochastic decoder and encoder.
By definition, $\mathcal K(z,\cdot)$ has the density $p_\theta(x|z)=\mathcal N(x;\mu_\theta(z),\Sigma_\theta(z))$ 
and $\mathcal R(x,\cdot)$ has the density $q_\phi(z|x)=\mathcal N(z;\mu_\phi(x),\Sigma_\phi(x))$.

Now, we aim to learn the parameters $\theta$ such that it holds approximately 
$$
p_X(x)\approx \int_{\R^n} p_\theta(x|z)p_Z(z)\dx z
$$
or equivalently
\begin{equation}
P_X(A)\approx\int_{\R^n}\mathcal K(z,A)\dx P_Z(z).\label{eq_VAE_integral}
\end{equation}
Assuming that the above equation holds true exactly, we can generate samples from $P_X$ by first sampling $z$ from $P_Z$ and then 
sampling $x$ from $\mathcal K(z,\cdot)$.

The first idea would be to use the maximum likelihood estimator as loss function, i.e., maximize
$$
\E_{x\sim P_X}[\log(p_\theta(x))], \qquad p_\theta(x)=\int_{\R^n} p_\theta(x|z)p_Z(z)\dx z.
$$
Unfortunately, computing the integral directly is intractable. 
Thus, using Bayes' formula
$$
p_\theta(z|x)=\frac{p_\theta(x|z)p_Z(z)}{p_\theta(x)},
$$
we artificially incorporate the stochastic encoder by the computation
\begin{align*}
\log(p_\theta(x))
&=\E_{z\sim q_\phi(\cdot|x)}\Big[\log\Big(\frac{p_\theta(x)p_\theta(z|x)}{p_\theta(z|x)}\Big)\Big]\\
&=\E_{z\sim q_\phi(\cdot|x)}\Big[\log\Big(\frac{p_\theta(x)p_\theta(z|x)}{q_\phi(z|x)}\Big)\Big]
+
\E_{z\sim q_\phi(\cdot|x)}\Big[\log\Big(\frac{q_\phi(z|x)}{p_\theta(z|x)}\Big)\Big]\\
&=
\E_{z\sim q_\phi(\cdot|x)}\Big[\log\Big(\frac{p_\theta(x|z)p_Z(z)}{q_\phi(z|x)}\Big)\Big]
+\mathrm{KL}(q_\phi(\cdot|x),p_\theta(\cdot|x))\\
&
\geq 
\E_{z\sim q_\phi(\cdot|x)}\Big[\log\Big(\frac{p_\theta(x|z)p_Z(z)}{q_\phi(z|x)}\Big)\Big] 
\end{align*}
Then the loss function 
\begin{equation}\label{eq_VAE_ELBO}
\mathcal L_{\theta,\phi}(x)\coloneqq \E_{z\sim q_\phi(\cdot|x)}\big[\log(p_\theta(x|z)p_Z(z))-\log(q_\phi(z|x)\big]
\end{equation}
is a lower bound on the so-called evidence $\log(p_\theta(x))$. 
Therefore it is called the \emph{evidence lower bound (ELBO)}.
Now the parameters $\theta$ and $\phi$ of the VAE $(E_\phi,D_\theta)$ 
can be trained by maximizing the expected ELBO, 
i.e., by minimizing the loss function
\begin{equation}\label{eq_VAE_loss}
\mathcal L_\text{VAE}(\theta,\phi)=-\E_{x\sim P_X}[\mathcal L_{\theta,\phi}(x)].
\end{equation}

\paragraph{VAEs as One Layer SNFs}

In the following, we show that a VAE is a special case of one layer SNF.
Let $\big((X_0,X_1),(Y_1,Y_0)\big)$ be a one-layer SNF, where the layers 
$\mathcal K_1=\mathcal K = P_{X_1|Z}$ and $\mathcal R_1=\mathcal R = P_{Y_0|X}$ 
are defined as in \eqref{eq_VAE_kernels} with
densities $p_\theta(\cdot|z)$ and $q_\phi(\cdot|x)$, respectively.
Note that in contrast to the stochastic layers from Section \ref{sec:snf} the dimensions $d_0$ and $d_1$ are no longer equal.
Now, with $X_0 = Z$, the loss function \eqref{eq_loss_SNF} of the SNF reads as
\begin{equation}\label{eq_VAE_SNF_loss}
\mathcal L_{\text{SNF}}(\theta,\phi)=\E_{(z,x)\sim P_{(Y_0,Y_1)}}\Big[-\log\Big(\frac{p_{X_1}(x)}{p_X(x) f_1(z,x)}\Big)\Big],
\end{equation}
where $f_1(\cdot,x)$ is given by the Radon-Nikodym derivative $\frac{\dx P_{Y_0|Y_1=x}}{\dx P_{X_0|X_1=x}}$. 
Now we can use that by the definition of $\mathcal K$ and $\mathcal R$ the random variables $(Y_0,Y_1)$ 
as well as the random variables $(X_0,X_1)$ have a joint density. 
Hence $f_1$ can be expressed by the corresponding densities $p_{X_0|X_1=x}$ of $p_{Y_0|Y_1=x}$. 
Together with  Bayes' formula we obtain
\begin{align}\label{logdet_VAE}
f_1(z,x)=\frac{p_{Y_0|Y_1=x}(z)}{p_{X_0|X_1=x}(z)}
=
\frac{q_\phi(z|x)}{p_{X_1|X_0=z}(x)}\frac{p_{X_1}(x)}{p_{X_0}(z)}=\frac{q_\phi(z|x)}{p_\theta(x|z)}\frac{p_{X_1}(x)}{p_{Z}(z)}.
\end{align}
Inserting this into \eqref{eq_VAE_SNF_loss}, we get
\begin{align}
\mathcal L_{\text{SNF}}(\theta,\phi)
&=\E_{(z,x)\sim P_{(Y_0,Y_1)}}\Big[-\log\Big(\frac{p_\theta(x|z)p_Z(z)}{q_\phi(z|x)p_X(x)}\Big)\Big]
\end{align}
and using \eqref{sense} further
\begin{align}
\mathcal L_{\text{SNF}}(\theta,\phi)
&=\E_{x\sim P_{X}}\Big[\E_{z\sim \mathcal R(x,\cdot)}\Big[-\log\Big(\frac{p_\theta(x|z)p_Z(z)}{q_\phi(z|x)p_X(x)}\Big)\Big]\Big]\\
&=\E_{x\sim P_{X}}\Big[\E_{z\sim q_\phi(\cdot,x)}\Big[-\log\Big(\frac{p_\theta(x|z)p_Z(z)}{q_\phi(z|x)}\Big)\Big]\Big]+\E_{x\sim P_X}[\log(p_X(x))]\\
&=-\E_{x\sim P_{X}}\Big[\mathcal L_{\theta,\phi}(x)\Big]+\E_{x\sim P_X}[\log(p_X(x))],
\end{align}
where $\mathcal L_{\theta,\phi}$ denotes the ELBO as defined in \eqref{eq_VAE_ELBO} and $\E_{x\sim P_X}[\log(p_X(x))]$ is a constant
independent of $\theta$ and $\phi$.
Consequently, minimizing $\mathcal L_{\text{SNF}}(\theta,\phi)$ 
is equivalent to minimize the negative expected ELBO, 
which is exactly the loss $\mathcal L_{\text{VAE}}(\theta,\phi)$ for VAEs from \eqref{eq_VAE_loss}.

The above result could alternatively be derived 
via the relation of the ELBO to the the KL divergence between the probability measures 
defined by the densities $(x,z)\mapsto p_\theta(x|z)p_Z(z)$ and $(x,z)\mapsto q_\phi(z|x)p_X(x)$, 
see \cite[Section 2.7]{Kingma_2019}.

\begin{remark}
Using the above result, we obtain by applying VAE layers within SNFs a natural way to combine VAEs with normalizing
flows, stochastic sampling methods and diffusion models.
Such models are widely used in the literature and have shown great performance. For example the authors of \cite{DJHWW2020, thin21,VKK2021} use a combination of VAE, MCMC and diffusion models. However,
each of these papers comes with its own analysis and derivation of the loss function. Using the SNF formulation we can put 
them all in one general framework without requiring a seperate analysis for each of them.
\end{remark}

\subsection{Diffusion Normalizing Flow Layer}\label{sec:special_snfs}
Recently, the authors of \cite{SSKKEP2020} proposed to learn the drift $g_t\colon\R^d\to\R^d$, $t\in[0,S]$ and diffusion coefficient $h_t\in\R$ 
of a stochastic differential equation
\begin{equation}\label{eq_sde}
\dx X_t=g_t(X_t)\dx t + h_t \dx B_t
\end{equation}
with respect to the Brownian motion $B_t$, such that it holds approximately $X_S\sim P_X$ for some $S>0$ and some data distribution $P_X$. 
The explicit Euler discretization of \eqref{eq_sde} with step size $\epsilon>0$ reads as
$$
X_t=X_{t-1}+\epsilon g_{t-1}(X_{t-1})+\sqrt{\epsilon} h_{t-1} \xi_{t-1}, \quad t=1,...,T,
$$
where $\xi_{t-1}\sim\mathcal N(0,I)$ is independent of $X_0,...,X_{t-1}$.
With a similar computation as for the Langevin layers, this corresponds to the Markov kernel 
\begin{equation}\label{eq_sde_forward_kernel}
\mathcal K_t(x,\cdot)= P_{X_t|X_{t-1}=x}=\mathcal N(x+\epsilon g_{t-1}(x),\epsilon h_{t-1}^2).
\end{equation}
Song et.\ al.\ parametrized the functions $g_t\colon\R^d\to\R^d$ by some a-priori learned score network, see \cite{HD2005,SE2019}, 
and achieved competitive performance in image generation.
Motivated by the time-reversal process \cite{A1982,HP1986,song2021maximum} of the SDE \eqref{eq_sde}, Zhang and Chen \cite{zhang2021diffusion} 
introduced the backward layer
$$
\mathcal R_t(x,\cdot)=P_{Y_{t-1}|Y_{t}=x} = \mathcal{N}(x+ \epsilon(g_t(x)-h_t^2 s_t(x)), \epsilon h_t^2)
$$
and learn the parameters of the neural networks $g_t\colon\R^d\to\R^d$ and $s_t\colon\R^d\to\R^d$ 
using the loss function $\mathcal L_{\text{SNF}}$ from \eqref{eq_loss_SNF} to achieve state-of-the-art results.
Even though Zhang and Chen call their model diffusion flow, it is indeed a special case of a SNF 
using the forward and backward layers $\mathcal K_t$ and $\mathcal R_t$.

On the other hand, not every SNF can be represented as a discretized SDE. For example, 
the forward layer \eqref{kernel_MCMC} from the MH layer does not have the form \eqref{eq_sde_forward_kernel}.

\subsection{Training of Stochastic Layers}\label{sec:stochastic_learning}
For the training of SNFs with the loss function $\mathcal L_\text{SNF}$ from \eqref{eq_loss_SNF}, we have to compute the quotients
$
\tfrac{f_t(x_{t-1},x_t)p_{X_{t-1}}(x_{t-1})}{p_{X_{t}}(x_{t})}
$
for every layer. The next theorem specifies, how this can be done for deterministic NF layers and the stochastic layers 
introduced in this section.

\begin{theorem}\label{lem_logdets}
Let $(X_0,...,X_T)$ be a Markov chain with a reverse $(Y_T,...,Y_0)$
and  $(x_0,...,x_T) \in \mathrm{supp}(P_{(X_0,...,X_T)})=\mathrm{supp}(P_{(Y_0,...,Y_T)})$. 
Let $f_t(\cdot,x_t)$ be the Radon-Nikodym derivative $\frac{\dx P_{Y_{t-1}|Y_t=x_t}}{\dx P_{X_{t-1}|X_t=x_t}}$.
Then the following holds true:
\begin{enumerate}
\item[\textrm{i)}] For the deterministic layer determined by \eqref{kern_det}:
$$
\frac{p_{X_{t-1}}(x_{t-1})}{p_{X_{t}}(x_{t})}=\frac1{|\nabla \mathcal T_t^{-1}(x_{t})|}
\quad\text{and}\quad 
f_t(x_{t-1},x_t)=1.
$$
\item[\textrm{ii)}]
For the Langevin layer in \eqref{eq_langevin_kernel}: 
$$
\frac{f_t(x_{t-1},x_t)p_{X_{t-1}}(x_{t-1})}{p_{X_{t}}(x_{t})}
=
\exp\Big(\frac12(\|\eta_t\|^2-\|\tilde\eta_t\|^2)\Big),
$$
where
$$
\eta_t\coloneqq \frac1{a_2}\big(x_{t-1}-x_t-a_1\nabla u_t(x_{t-1})\big),
\quad \tilde\eta_t\coloneqq \frac1{a_2}\big(x_{t-1}-x_t+a_1\nabla u_t(x_{t})\big).
$$
\item[\textrm{iii)}] For the MH layer determined by \eqref{kernel_MCMC}:
$$
\frac{f_t(x_{t-1},x_t)p_{X_{t-1}}(x_{t-1})}{p_{X_{t}}(x_{t})}=\frac{p_t(x_{t-1})}{p_t(x_{t})}.
$$
\item[\textrm{iv)}] For the MALA layer given by \eqref{kernel_MALA}:
$$
\frac{f_t(x_{t-1},x_t)p_{X_{t-1}}(x_{t-1})}{p_{X_{t}}(x_{t})}=\frac{p_t(x_{t-1})}{p_t(x_{t})}.
$$
\item[\textrm{v)}] For the VAE layer in \eqref{eq_VAE_kernels}:
$$
\frac{f_t(x_{t-1},x_t)p_{X_{t-1}}(x_{t-1})}{p_{X_{t}}(x_{t})}=\frac{q_\phi(x_{t-1}|x_t)}{p_\theta(x_t|x_{t-1})}
$$

\item[\textrm{vi)}] For the diffusion normalizing flow layer in \eqref{eq_sde_forward_kernel}:
$$
\frac{f_t(x_{t-1},x_t)p_{X_{t-1}}(x_{t-1})}{p_{X_{t}}(x_{t})}
=
\exp\Big(\frac12(\|\eta_t\|^2-\|\tilde\eta_t\|^2)\Big),
$$
where
\begin{align*}
\eta_t&\coloneqq \frac1{\sqrt{\epsilon}g_{t-1}}\big(x_{t-1}-x_t+\epsilon f_{t-1}(x_{t-1})\big),\\
\tilde\eta_t&\coloneqq \frac1{\sqrt{\epsilon}g_{t}}\big(x_{t-1}-x_t-\epsilon (f_t(x_t)-g_t^2s_t(x_t))\big).
\end{align*}
\end{enumerate}
\end{theorem}

The assertions i)-iv) were proved in \cite{HHS2021}, while v) follows from equation \eqref{logdet_VAE} and the derivation of vi)
is analogously to ii).

\section{Conditional Generative Modeling} \label{sec:conditional}
\begin{figure}
  \includegraphics[width=0.48\linewidth]{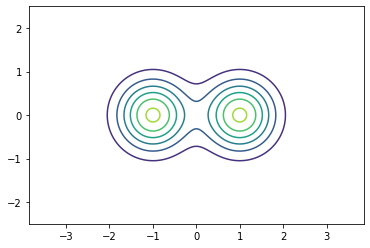}  
  \includegraphics[width=0.48\linewidth]{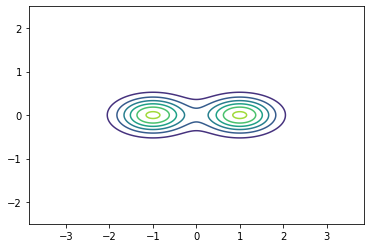} 
\caption{Illustration of the prior density $p_X$ (left) and the posterior density $p_{X|Y=y}$ for $y=0$ (right) within the inverse problem
\eqref{eq_inverse_problem} with $F(x_1,x_2)=x_2$ and $\eta\sim\mathcal N(0,0.1^2)$.}
\label{Bayes_fig}
\end{figure}

So far we have considered the task of sampling from $P_X$ using a simpler distribution $P_Z$.
For inverse problems, we have to adjust our setting.
Let $X\colon\Omega\to\R^d$ be a random variable and let $Y\colon\Omega\to\R^{\tilde d}$ be defined by
\begin{align}\label{eq_inverse_problem}
Y=F(X)+\eta,
\end{align}
for some (ill-posed), not necessary linear operator $F\colon\R^d\to\R^{\tilde d}$ 
and a random variable $\eta\colon\Omega\to\R^{\tilde d}$.
In the following, we aim to find the posterior distribution $P_{X|Y=y}$ for some measurement $y\in\R^{\tilde d}$.
In the case that $X$ and $Y$ have a joint density, this can be done by Bayes' theorem, which states that
$$
p_{X|Y=y}(x)=\frac{p_{Y|X=x}(y)p_X(x)}{p_Y(y)}.
$$
Under the assumption that $y$ is fixed, the term $p_Y(y)$ is just a constant, such that $p_{X|Y=y}$ is 
(up to a multiplicative constant) given by the product of the likelihood $p_{Y|X=x}(y)$ and the prior $p_X(x)$.
Figure~\ref{Bayes_fig} illustrates the prior density $p_X$ on the left and the posterior density $p_{X|Y=y}$ for $y=0$ on the right
for the inverse problem \eqref{eq_inverse_problem} with $F(x_1,x_2)=x_2$ and $\eta\sim\mathcal N(0,0.1^2)$.
It can be seen that the observation $Y=y$ modifies the density $p_X$ such that the distribution $p_{X|Y=y}$ is concentrated
around the pre-image $F^{-1}(\{y\})$.
In the literature, the problem of recovering the posterior distribution $P_{X|Y=y}$ was tackled by MCMC methods by \cite{robertsrosenthalMCMC} 
and conditional normalizing flows by \cite{AFHHSS2021,ALKRK2019, DSLM2021,WWHW2019} or conditional VAEs by \cite{SLY2015}.
In the following, we show similarly as before that conditional SNFs include all these methods.
We are also aware of the concept of conditional GANs by \cite{MO2014}. However, as we are not sure how they are related to SNFs, 
we do not consider them in detail.

\subsection{Conditional Normalizing Flows}

A conditional normalizing flow is a mapping $\mathcal T=\mathcal T_\theta\colon\R^{\tilde d}\times\R^{d}\to\R^d$ 
such that for any $y\in\R^{\tilde d}$, 
the function $\mathcal T(y,\cdot)$ is invertible.
Here, we call $y$ the \emph{condition} of $\mathcal T(y,\cdot)$.
Then, it can be used to model the density $p_{X|Y=y}$ of $P_{X|Y=y}$ for an arbitrary $y\in\R^{\tilde d}$ by a simpler distribution $P_Z$, 
by learning $\mathcal T$ such that it holds approximately
$$
P_{X|Y=y}\approx\mathcal T(y,\cdot)_\#P_Z,\quad\text{or equivalently}\quad P_Z\approx \mathcal T(y,\cdot)^{-1}_\# P_X.
$$
Similarly as in the non-conditional case this approximation can be done by the expected Kullback-Leibler (or expected maximum likelihood loss, see \cite{AKLBRK2021}) divergence
\begin{align}
&\quad\E_{y\sim P_Y}[\mathrm{KL}(P_{X|Y=y},\mathcal T(y,\cdot))]\\
&=\E_{y\sim P_Y}\Big[\E_{x\sim P_{X|Y=y}}\Big[\log\Big(\frac{p_{X|Y=y}(x)}{p_{\mathcal T(y,\cdot)_\#P_Z}(x)}\Big)\Big]\Big]\\
&=\E_{(x,y)\sim P_{X,Y}}[p_{X|Y=y}]-\E_{(x,y)\sim P_{X,Y}}\Big[\log \big( p_Z\circ \mathcal T^{-1}(y,x) \big)\\
&\quad
+ \log|\mathrm{det}(\nabla\mathcal T^{-1}(y,x))|\Big],
\end{align}
where $\mathcal T^{-1}(y,x)$ is the function $\mathcal T^{-1}(y,\cdot)$ evaluated at $x$ and $\nabla \mathcal T^{-1}(y,x)$ denotes 
its Jacobian.
As the first summand is constant, we obtain the loss function
$$
\mathcal L_\text{cNF}(\theta)\coloneqq
-\E_{(x,y)\sim P_{X,Y}}\Big[\log \big( p_Z\circ \mathcal T^{-1}(y,x) \big) + \log|\mathrm{det}(\nabla\mathcal T^{-1}(y,x))|\Big].
$$
Note that this derivation needs the expected value over many observations $y$, because the forward Kullback--Leibler divergence would otherwise require samples from the posterior $P_{X|Y=y}$. However by averaging over many observations, we get that we "only" need to sample from the joint distribution (which is available if one knows the prior and the forward model).
Now, let $\mathcal T$ be the composition of multiple blocks, i.e.,
$$
\mathcal T(y,\cdot)=\mathcal T_T(y,\cdot)\circ\cdots\circ \mathcal T_1(y,\cdot).
$$
Then, the blocks generate two sequences $(X_0,...,X_T)$ and $(Y_T,...,Y_0)$ of random variables
$$
X_0\sim P_Z,\quad X_t=\mathcal T_t(Y,X_{t-1})\qquad \text{and}\qquad Y_T=X,\quad Y_{t-1}=\mathcal T_t^{-1}(Y,Y_t).
$$
Due to the condition, it is now intractable to compute the kernels $P_{X_t|X_{t-1}}$ and $P_{Y_{t-1}|Y_t}$. 
Instead, we consider the kernels $\mathcal K_t=P_{X_t|X_{t-1},Y}$ and $\mathcal R_t=P_{Y_{t-1}|Y_t,Y}$. 
By a similar computation as in the non-conditional case, they are given by
$$
\mathcal K_t(y,x,\cdot)=\delta_{\mathcal T_t(y,x)}\quad \text{and}\quad \mathcal R_t(y,x,\cdot)=\delta_{\mathcal T_t^{-1}(y,x)}.
$$
Note, that for an arbitrary $y$ the distributions $P_{(X_0,...,X_T)|Y=y}$ and $P_{(Y_T,...,Y_0)|Y=y}$ are determined by
$$
P_Z\times \mathcal K_1(y,\cdot,\cdot)\times\cdots\times \mathcal K_T(y,\cdot,\cdot)
\quad\text{and}\quad 
P_{X|Y=y}\times \mathcal R_T(y,\cdot,\cdot)\times\cdots\times \mathcal R_1(y,\cdot,\cdot)
$$
such that any two sequences $(X_0^y,...,X_T^y)$ and $(Y_T^y,...,Y_0^y)$ following these distributions are Markov chains.

\begin{remark}
Furthermore, similar to the unconditional case we can also consider the backward KL, cf.~Remark~\ref{rem_backward_KL}. 
This would then give
\begin{align}
	&\quad\E_{y\sim P_Y}[\mathrm{KL}(T(y,\cdot)_\#P_Z,P_{X|Y=y})]\\
	&=\E_{y\sim P_Y}[\mathrm{KL}(P_Z,T(y,\cdot)^{-1}_\#P_{X|Y=y})]\\
	&=\E_{y,z}[-\log p_{X|Y=y}(T(y,z))-\log|\mathrm{det}(\nabla\mathcal T(y,z))]\\
	&=\E_{y,z}[-\log p_{Y|X=T(y,z)}(y) - \log p_X(T(y,z))+ \log p_Y(y)\\
	&\qquad\qquad-\log|\mathrm{det}(\nabla\mathcal T(y,z))]
\end{align}
where we can drop the evidence $p_Y(y)$ in the optimization with respect to $\mathcal{T}$. This formulation has been used in several papers, see \cite{AH2022,AFHHSS2021,KDSK2020,Sun2021DeepPI}.
\end{remark}

\subsection{Conditional SNFs}

As in the non-conditional case, we obtain conditional SNFs from conditional normalizing flows 
by replacing some of the deterministic transitions $\mathcal T_t$ by random transforms. 
In terms of Markov kernels, we replace the Dirac measures from $\mathcal K_t$ and $\mathcal R_t$ by more general kernels.
This leads to the following formal definition of conditional SNFs.
A \emph{conditional SNF} is a pair of sequences $((X_0,...,X_T),(Y_T,...,Y_0))$ of random variables $X_t,Y_t\colon\Omega\to\R^{d_t}$ such that:
\begin{itemize}
\item[] cP1) the conditional distributions $P_{X_t|Y=y}$ and $P_{Y_t|Y=y}$ have densities 
$$
p_{X_t}(y,\cdot)\colon\R^{d_t}\to\R_{>0},\quad\text{and}\quad p_{Y_t}(y,\cdot)\colon\R^{d_t}\to\R_{>0}
$$
for $P_Y$-almost every $y$ and
all $t=1,\ldots,T$,
\item[] cP2) for $P_Y$-almost every $y$, there exist Markov kernels $\mathcal K_t\colon\R^{\tilde d}\times\R^{d_{t-1}}\times \mathcal B(\R^{d_t})\to[0,1]$ 
and $\mathcal R_t\colon\R^{\tilde d}\times \R^{d_t}\times \mathcal B(\R^{d_{t-1}})\to[0,1]$
such that
\begin{align}
P_{(X_0,...,X_T)|Y=y}=P_{X_0} \times \mathcal K_1(y,\cdot,\cdot)\times\cdots\times \mathcal K_T(y,\cdot,\cdot),\\
P_{(Y_T,...,Y_0)|Y=y}=P_{Y_0} \times \mathcal R_T(y,\cdot,\cdot)\times\cdots\times \mathcal R_1(y,\cdot,\cdot).
\end{align}
\item[] cP3) for $P_{Y,X_t}$-almost every pair $(y,x)\in\R^{\tilde d}\times\R^{d_t}$,
the measures $\mathcal R_t(y,x,\cdot)$ and  $P_{X_{t-1}|X_t=x,Y=y}(\cdot)$
are absolute continuous with respect to each other.
\end{itemize}
We call the sequence $(Y_T,...,Y_0)$ a \emph{reverse} of $(X_0,...,X_T)$.
For applications, one usually sets
$$
X_0=Z,
$$
where $Z$ is a random variable, which is easy to sample from and we aim to approximate 
for any $y\in\R^{\tilde d}$ the distribution $P_{X|Y=y}$ by $P_{X_T|Y=y}$. On the other hand, the reverse usually starts with 
$$
P_{Y_T|Y=y}=P_{X|Y=y}
$$
and $P_{Y_0}$ should approximate the latent distribution $P_Z$.

The stochastic layers can be chosen analogously as in the non-conditional case. For details, we refer to \cite{HHS2021}.

For training conditional SNFs, we aim to minimize the Kullback-Leibler divergence
$$
\mathrm{KL}(P_{Y,(Y_0,...,Y_T)},P_{Y,(X_0,...,X_T)}).
$$
By \cite[Corollary 9]{HHS2021}, this is equivalent to minimizing
\begin{align*}
\mathcal L_{\text{cSNF}}(\theta)&=\E_{y\sim P_Y}\Big[\E_{(x_0,...,x_T)\sim P_{Y_0^y,...,Y_T^y}}\Big[ \\
& \quad \log\Big(\frac{1}{p_Z(x_0)}\prod_{t=1}^T\frac{f_{t,y}(x_{t-1},x_t)p_{X_{t-1}^y}(x_{t-1})}{p_{X_t^y}(x_t)}\Big)\Big]\Big],
\end{align*}
where $f_{t,y}(\cdot,x_t)$ is given by the Radon-Nikodym derivative $\frac{\dx P_{Y_{t-1}^y|Y_t^y=x_t}}{\dx P_{X_{t-1}^y|X_t^y=x_t}}$ 
and where the pair $((X_0^y,...,X_T^y),(Y_T^y,...,Y_0^y))$ of Markov chains follows the distributions
$$
P_Z\times \mathcal K_1(y,\cdot,\cdot)\times\cdots\times \mathcal K_T(y,\cdot,\cdot)
\quad\text{and}\quad 
P_X\times \mathcal R_T(y,\cdot,\cdot)\times\cdots\times \mathcal R_1(y,\cdot,\cdot).
$$
\begin{remark}
Once again, we can formulate the backward KL loss function for the conditional SNF case, namely via 
\begin{align*}
\tilde{\mathcal L}_{\text{cSNF}}(\theta)&=\E_{y\sim P_Y}\Big[\E_{(x_0,...,x_T)\sim P_{X_0^y,...,X_T^y}}\Big[ \\
& \quad \log\Big(\frac{1}{p_{X|Y=y}(x_T)}\Big) - \log\Big(\prod_{t=1}^T\frac{f_{t,y}(x_{t-1},x_t)p_{X_{t-1}^y}(x_{t-1})}{p_{X_t^y}(x_t)}\Big)\Big]\Big].
\end{align*}
\end{remark}
\begin{figure}
\begin{center}
\includegraphics[width=.95\textwidth]{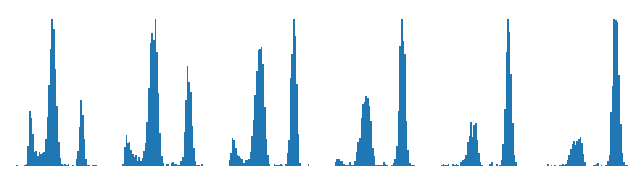}
\end{center}
\caption{One dimensional marginals of the interpolations $P_{X|Y=y}$ with $y=(1-\lambda)y_0+\lambda y_1$ between 
the posterior distributions $P_{X|Y=y_0}$ and $P_{X|Y=y_1}$ using equidistant $\lambda\in[0,1]$ for the 
inverse problem from Section~\ref{sec:scatterometry}.}
\label{latent_manipulation}
\end{figure}

\begin{example}
Once a conditional (stochastic) normalizing flow is trained, we can reconstruct the posterior distribution $P_{X|Y=y}$ for any observation
$y$. In particular, this allows us to interpolate between the posterior distributions $P_{X|Y=y_0}$ and $P_{X|Y=y_1}$ by the distributions
$P_{X|Y=y}$ for $y=(1-\lambda)y_0+\lambda y_1$.
We plot the one dimensional marginals of such an interpolation for the example from Section~\ref{sec:scatterometry} in 
Figure~\ref{latent_manipulation}.
\end{example}

\subsection{Conditional VAEs}

\cite{SLY2015} proposed conditional VAEs. A \emph{conditional VAE} is a pair $(E,D)$ of a conditional stochastic decoder
$$
D(y,z)=D_\theta(y,z)\coloneqq(\mu_\theta(y,z),\Sigma_\theta(y,z))
$$
and a conditional stochastic encoder
$$
E(y,x)=E_\phi(y,x)\coloneqq(\mu_\phi(y,x),\Sigma_\phi(y,x)).
$$
The networks $E$ and $D$ are learned such that the kernels
\begin{align*}
\mathcal K(y,z,\cdot)&\coloneqq \mathcal N(\mu_\theta(y,z),\Sigma_\theta(y,z))
,\\
\mathcal R(y,x,\cdot)&\coloneqq \mathcal N(\mu_\phi(y,x,\cdot),\Sigma_\phi(y,s,\cdot))
\end{align*}
push the probability distribution $P_Z$ onto $P_{X|Y=y}$ and vice versa. As a loss function, a straight forward modification of
the ELBO \eqref{eq_VAE_ELBO} is used.
Similar, as in the non-conditional case, it turns out that a conditional VAE is a one-layer conditional SNF with layers $\mathcal K$
and $\mathcal R$ as defined above.

\section{Numerical Results} \label{sec:numerics}
%
In this section, we present three numerical examples of applications of conditional SNFs to inverse problems and an artificial example related to dynamic optimal transport.
The first one with Gaussian mixture models is academical, but quite useful, 
since we know the ground truth posterior distribution, which we aim to approximate.
This enables us to give quantitative comparisons.
The second example is a real world example from scatterometry. 
Both setups were also used in \cite{HHS2021}, but the models and comparisons are different.
The code for the numerical examples is available online\footnote{\url{https://github.com/PaulLyonel/Gen_norm_flow}}.

The last one is an example of image generation as 2d densities, where we interpolated two images using optimal transport\footnote{For generating the energies and samples from an image we use the code from \cite{WKN2020}.}.

\begin{remark}[Proposal densities for inverse problems]
For using Langevin layers, MH layers or MALA layers we need again a proposal density.
As in the unconditional case, this proposal density interpolates between the target density $p_{X|Y=y}$ and 
the latent density $p_Z$. The target density can up to a constant be rewritten by Bayes theorem as
$p_{X|Y=y}(x)\propto p_X(x)p_{Y|X=x}(y)$. Thus, the computation of the proposal density consists of the computation
of the prior and the likelihood. The prior is either assumed to be known or can be approximated accordingly to Remark~\ref{rem_prop_den}.
The likelihood computation consists of the forward operator and the noise model.
\end{remark}

\subsection{Posterior Approximation for Gaussian Mixtures }\label{sec:mm}
To verify that our proposed methods yield the correct posteriors, 
we apply our framework to a linear inverse problem with a Gaussian mixture model, 
where we can analytically infer the ground truth posterior distribution
by the following lemma, which can also be found in \cite{HHS2021}.

\begin{lemma} \label{mm}
Let $X \sim \sum_{k=1}^K w_k \mathcal N(m_k,\Sigma_k)$.
Suppose that 
$$
Y=AX+\eta,
$$
where
$A: \R^d \rightarrow \R^{\tilde d}$ 
is a linear operator and we have Gaussian noise 
$\eta \sim N(0,b^2 I)$. Then   
$$
P_{X|Y=y} \propto \sum_{k=1}^K \tilde w_k \mathcal N(\cdot|\tilde m_k,\tilde \Sigma_k),
$$
where $\propto$ denotes equality up to a multiplicative constant and
$$
\tilde \Sigma_k \coloneqq (\tfrac{1}{b^2}A^\tT A+\Sigma_k^{-1})^{-1},
\qquad 
\tilde m_k \coloneqq \tilde\Sigma_k (\tfrac1{b^2}A^\tT y+\Sigma_k^{-1} m_k).
$$
and
$$
\tilde w_k \coloneqq \frac{w_k}{|\Sigma_k|^{\tfrac12}} \exp\left(\frac12 (\tilde m_k \tilde \Sigma_k^{-1} \tilde m_k - m_k \Sigma_k^{-1} m_k)\right).
$$
\end{lemma}

\begin{proof}
First, consider one component  $P_{X_k} = \mathcal N(m_k,\Sigma_k)$.
Using Bayes' theorem we get
\begin{align*}
p_{X_k|Y_k=y}(x) 
&= p_{Y_k|X_k=x}(y)p_{X_k}(x)/p_{Y_k}(y), \\
p_{Y_k}(y) &= \int_{\R^{\tilde d}} p_{X_k}(z) p_{Y_k|X_k = z}(y) \dx z
\end{align*}
and further
\begin{align*}
&p_{Y_k|X_k=x}(y) p_{X_k}(x)
=
\mathcal N(y|Ax, b^2 I) \mathcal N(x|m_k, \Sigma_k)\\
&\propto
\frac{1}{|\Sigma_k|^\frac12}
\exp\left( -\tfrac12 (x-\tilde m_k)^\tT \tilde \Sigma^{-1} (x- \tilde m_k)  \right)
\exp\left( - \tfrac12  m_k^\tT \Sigma_k^{-1}  m_k 
+ \tfrac12  \tilde m_k^\tT \tilde \Sigma_k^{-1}  \tilde m_k \right)
\end{align*}
with a constant independent of $k$.
Then we get for  the mixture model
$P_X  \sim \sum_{k=1}^K w_k \mathcal N(m_k,\Sigma_k)$
again by Bayes' theorem that
\begin{align*}
&p_{X|Y=y}(x)\propto\sum_{k=1}^K w_k p_{Y|X=x}(y)\mathcal N(x|m_k,\Sigma_k)\\
&\propto 
\sum_{k=1}^K \frac{w_k}{|\Sigma_k|^\frac12}
\exp\left( -\tfrac12 (x-\tilde m_k)^\tT \tilde \Sigma^{-1} (x- \tilde m_k)  \right)
\exp\left( - \tfrac12  m_k^\tT \Sigma_k^{-1}  m_k 
+ \tfrac12  \tilde m_k^\tT \tilde \Sigma_k^{-1}  \tilde m_k \right)\\
&\propto 
 \sum_{k=1}^K \tilde w_k \mathcal N(\cdot|\tilde m_k,\tilde \Sigma_k).
\end{align*}
\end{proof}

In our experiment, we consider the Gaussian mixture model on $\mathbb R^{100}$ given by
$$
X \sim \sum_{k=1}^5 \frac15 \mathcal N(m_k,\Sigma_k), \quad \Sigma_k = 10^{-4} \, I,
$$
where the means were chosen uniformly in the interval $[-1,1]$.
As operator $A$ we use the diagonal matrix with entries $a_{i,i} = 0.1/(i+1)^2$.
Finally, we choose $b = 0.05$ for the Gaussian noise $\eta$. 
Note that the variance of the mixture components was chosen very small, 
so that the posterior is very concentrated around its modes. 
This makes the problem particularly challenging and 
underpins the need for stochastic layers. 
Furthermore, the dimension $100$ of this problem is quite large for Bayesian inversion.  

We test the following 7 conditional models, which all have a similar number of parameters ranging from 779840 to 882240.
All were trained for 5000 iterations, where the pure VAE model trained the quickest. The INN MALA model was slower by a factor of 2, and the full flow model was the slowest, but we included it for comparison in the evaluation as it learns the geometric annealing schedule. 

Here we give a more accurate description of the used models. When we speak of VAEs we mean layers of the form $\mathcal{K}(x,y,\cdot) \sim \mathcal{N}(f(x,y), \sigma^2)$ and $\mathcal{R}(x,y,\cdot) \sim \mathcal{N}(g(x,y), \sigma^2)$, i.e. we do not learn the covariances. 
Note that the quotient in the loss for the VAE layer in Lemma \ref{lem_logdets} penalizes if the networks $f$ and $g$ are not "inverses" to each other.

\begin{itemize}
\item \textbf{INN}: Here we took 8 coupling layers with a subnetwork with 2 hidden layers containing 128 neurons in them. 
\item \textbf{VAE}: They consist of a forward and backward ReLU neural network with 2 hidden layers of size 128. We concatenated 8 layers of them, with noise levels of $0.1$ and a noise level of $0.01$ in the final layer. 
\item \textbf{INN and VAE}: Here we took 4 layers, which consisted of one coupling and one autoencoder as above. Furthermore, in the final layer we used noise level $0.01$.
\item \textbf{MALA and INN}: Same architecture as the conditional INN and a MALA layer in the last layer. The MALA layer uses 3 Metropolis--Hastings steps with step size of 5e-5. 
\item \textbf{MALA and VAE}: Same architecture as the autoencoder (8 layers with noise level of $0.1$) with a MALA layer as the final layer. 
\item \textbf{MALA and VAE and INN}: Same architecture as the conditional INN and autoencoder, but with a MALA layer as the final layer.
\item \textbf{FULL FLOW}: Here we use the geometric annealing schedule between latent distribution and posterior. For the first 4 layers we alternate between conditional INN layers and then alternate between 4 conditional VAE layers and MALA layers. Each MALA layer uses 5 Metropolis steps and step size of 5e-3. For the last layer we used 5e-5.
\end{itemize}

\begin{figure}
\begin{subfigure}{0.49\textwidth}
  \centering
  \includegraphics[width=\textwidth]{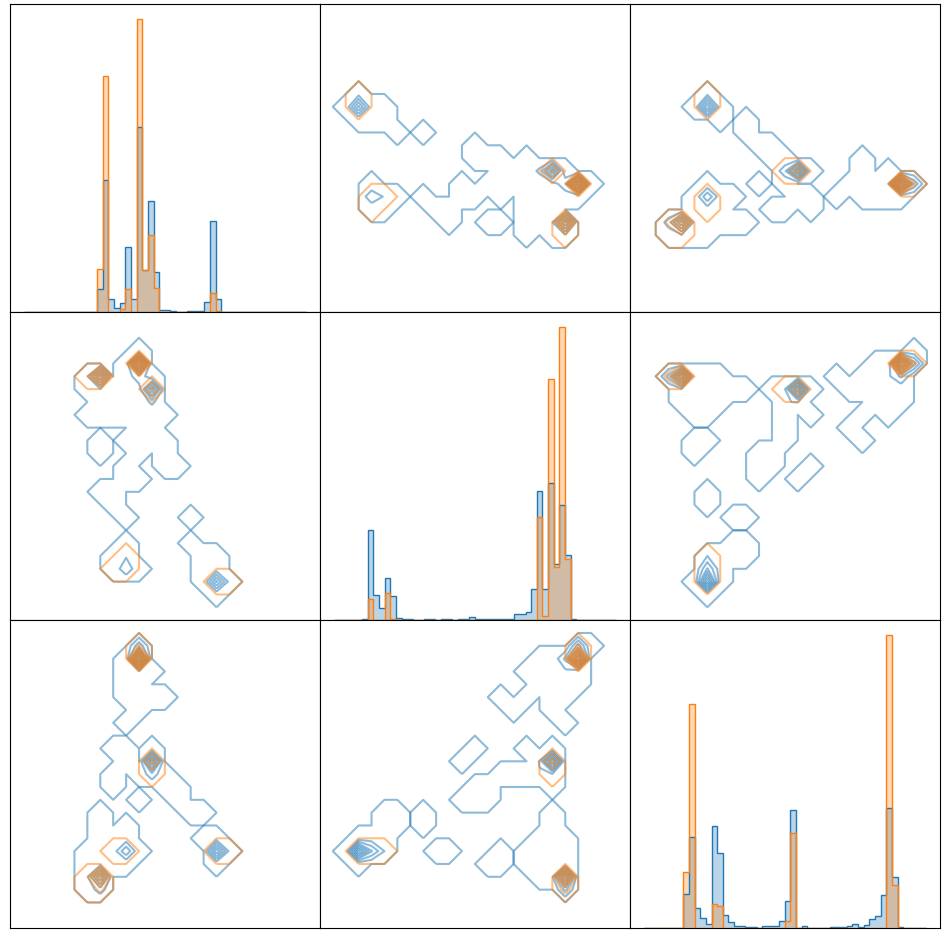}  
  \vspace{-.6cm}
  \caption*{INN}
\end{subfigure}
\begin{subfigure}{0.49\textwidth}
  \centering
	\includegraphics[width=\textwidth]{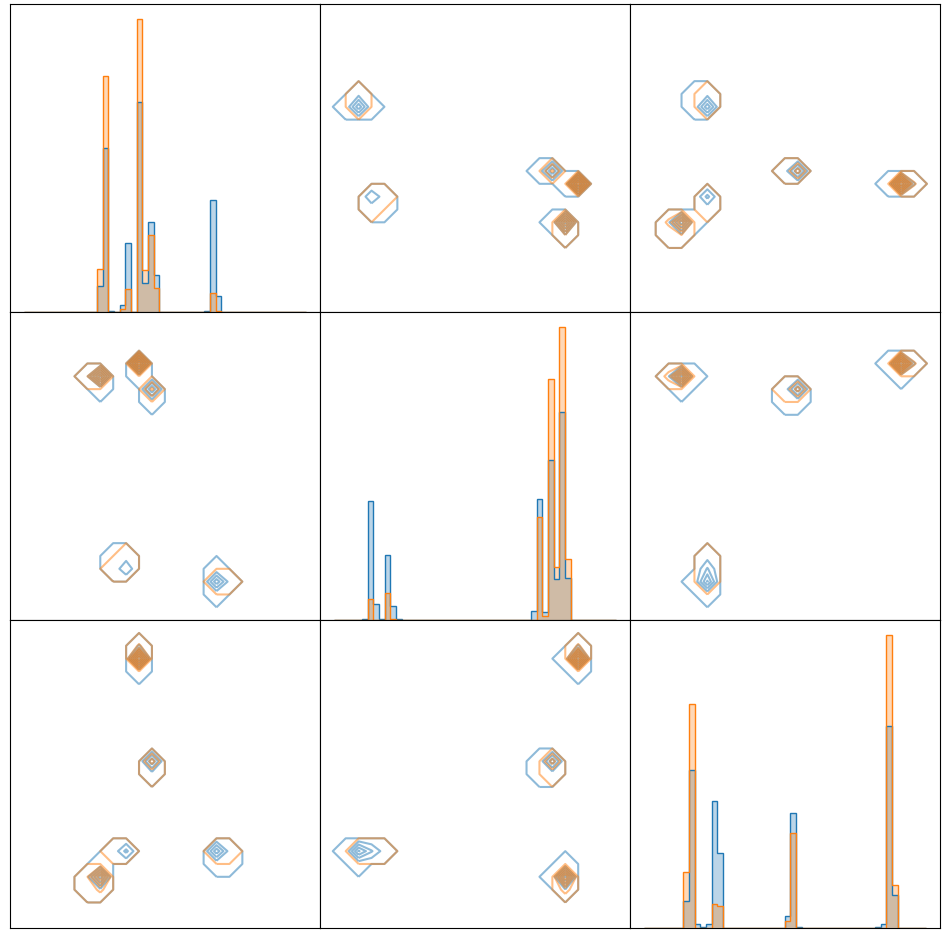}  
  \vspace{-.6cm}
  \caption*{INN + MALA}
\end{subfigure}
	
\begin{subfigure}{0.49\textwidth}
  \centering
  \includegraphics[width=\textwidth]{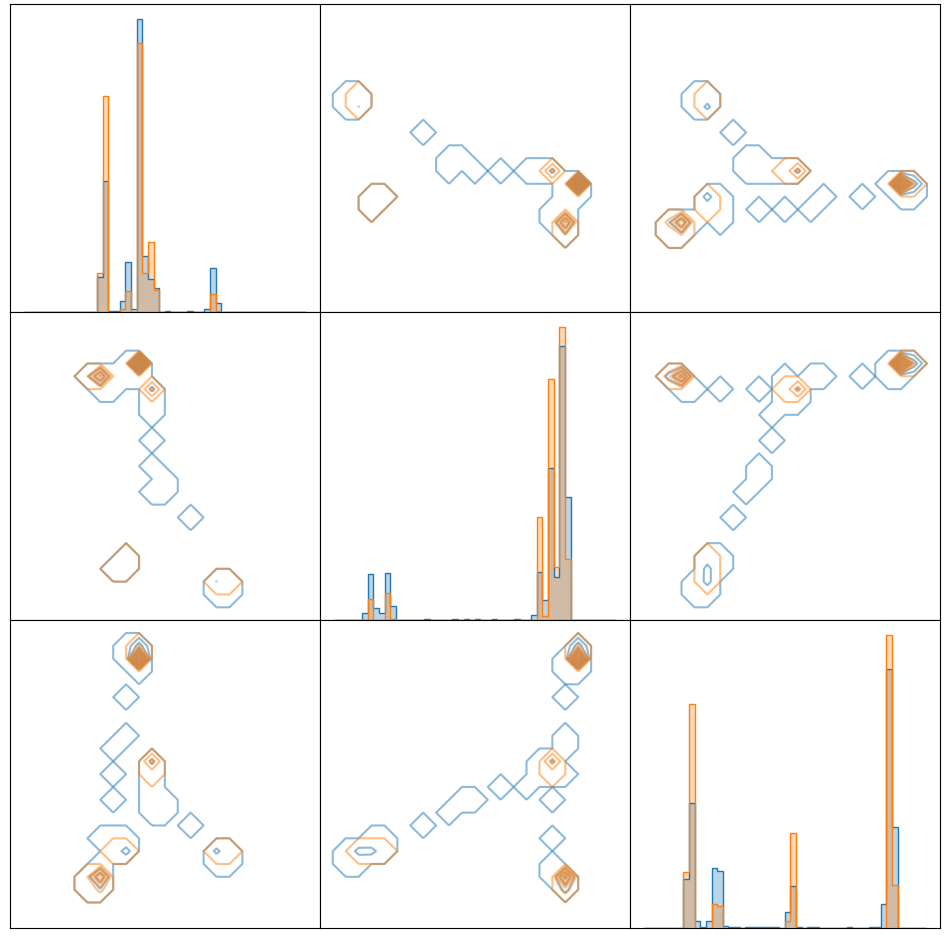}  
  \vspace{-.6cm}
  \caption*{VAE}
\end{subfigure}
\begin{subfigure}{0.49\textwidth}
  \centering
  \includegraphics[width=\textwidth]{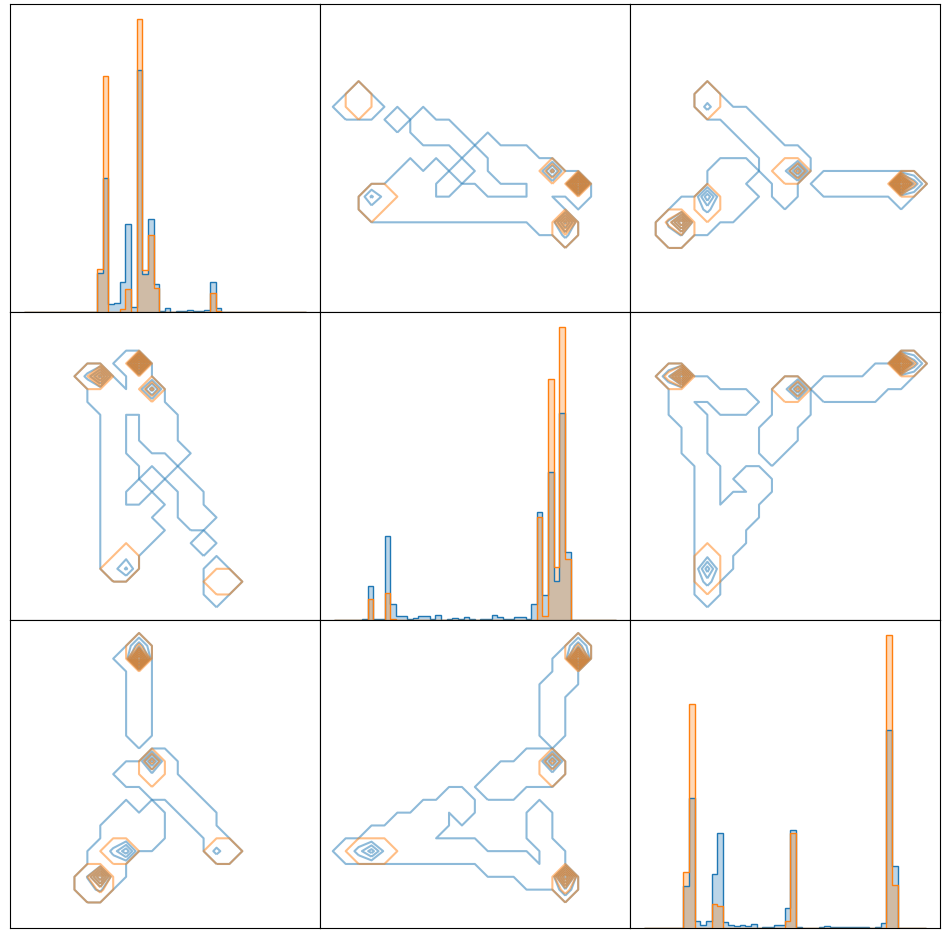}  
  \vspace{-.6cm}
  \caption*{INN + VAE}
\end{subfigure}

\begin{subfigure}{0.49\textwidth}
  \centering
  \includegraphics[width=\textwidth]{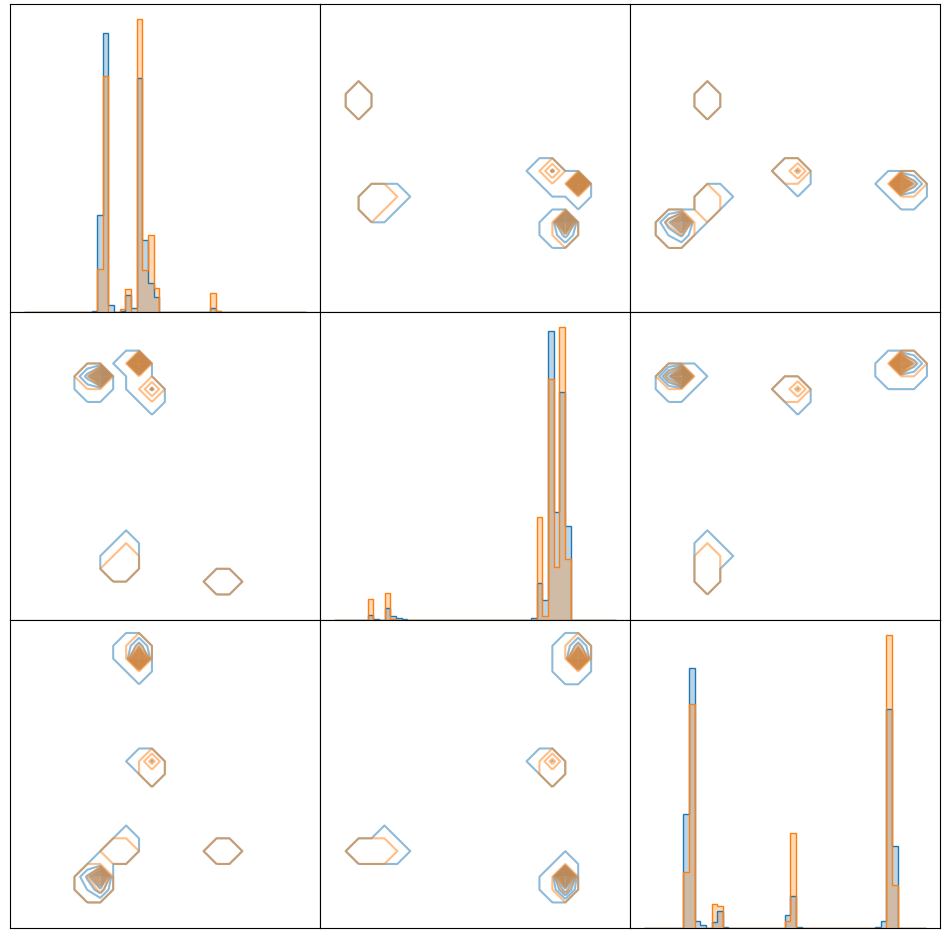}  
  \vspace{-.6cm}
  \caption*{VAE + MALA}
\end{subfigure}
\begin{subfigure}{0.49\textwidth}
  \centering
  \includegraphics[width=\textwidth]{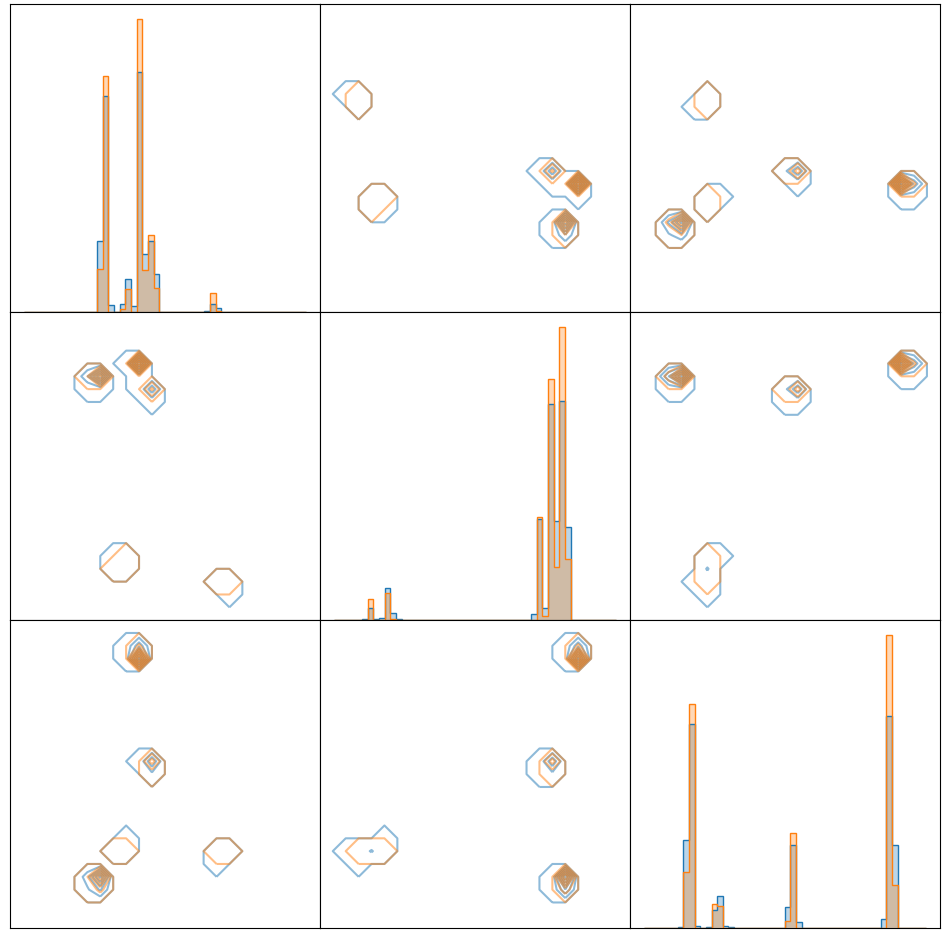}  
  \vspace{-.6cm}
  \caption*{INN+VAE+MALA}
\end{subfigure}
\caption{Histograms and 2-marginals of the first, 
$50$-th and $100$-th marginal of the ground truth posterior distribution (orange) and the posterior reconstructions of the six
different models. 
}
\label{fig:mix_models}
\end{figure}

\begin{figure}
\begin{subfigure}{0.49\textwidth}
  \centering
  \includegraphics[width=\textwidth]{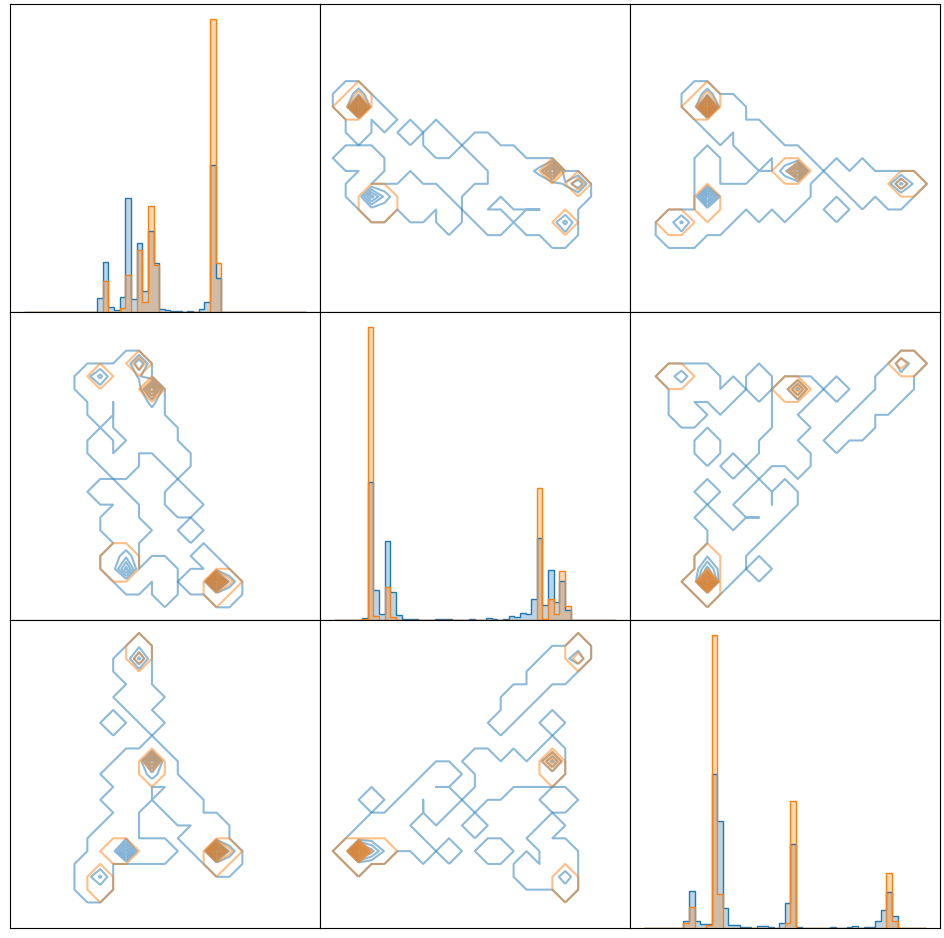}  
  \vspace{-.6cm}
  \caption*{INN}
\end{subfigure}
\begin{subfigure}{0.49\textwidth}
  \centering
	\includegraphics[width=\textwidth]{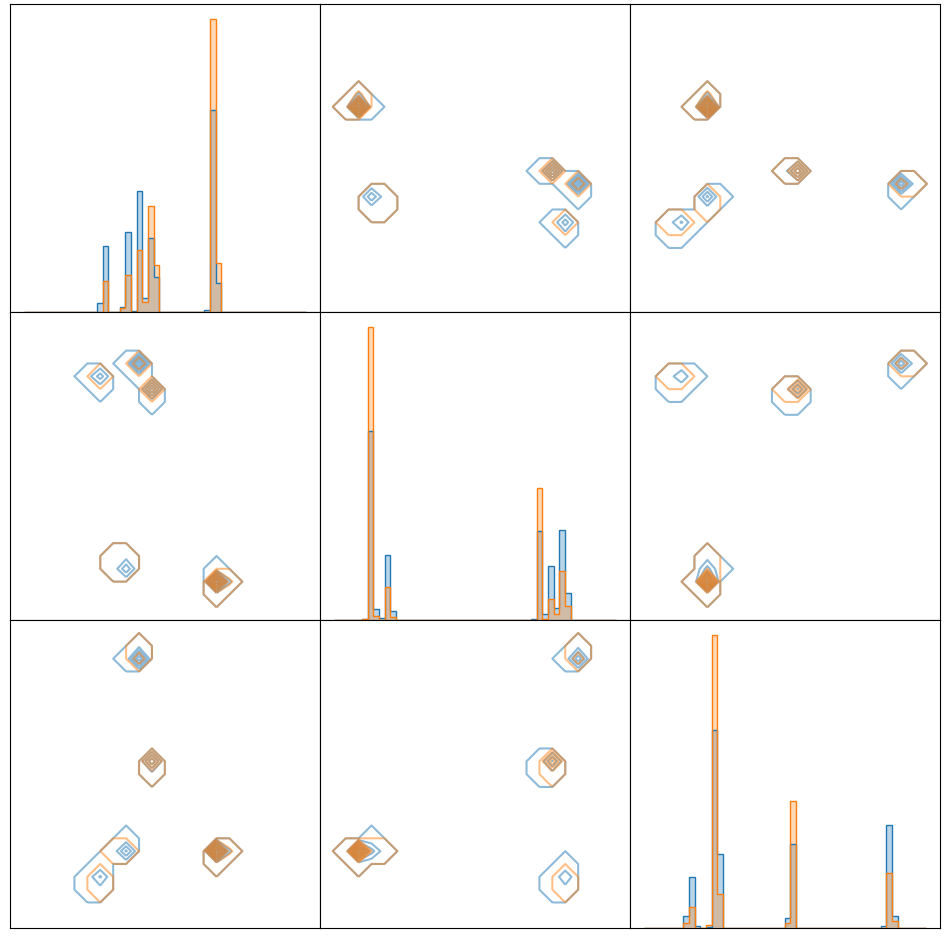}  
  \vspace{-.6cm}
  \caption*{INN + MALA}
\end{subfigure}
	
\begin{subfigure}{0.49\textwidth}
  \centering
  \includegraphics[width=\textwidth]{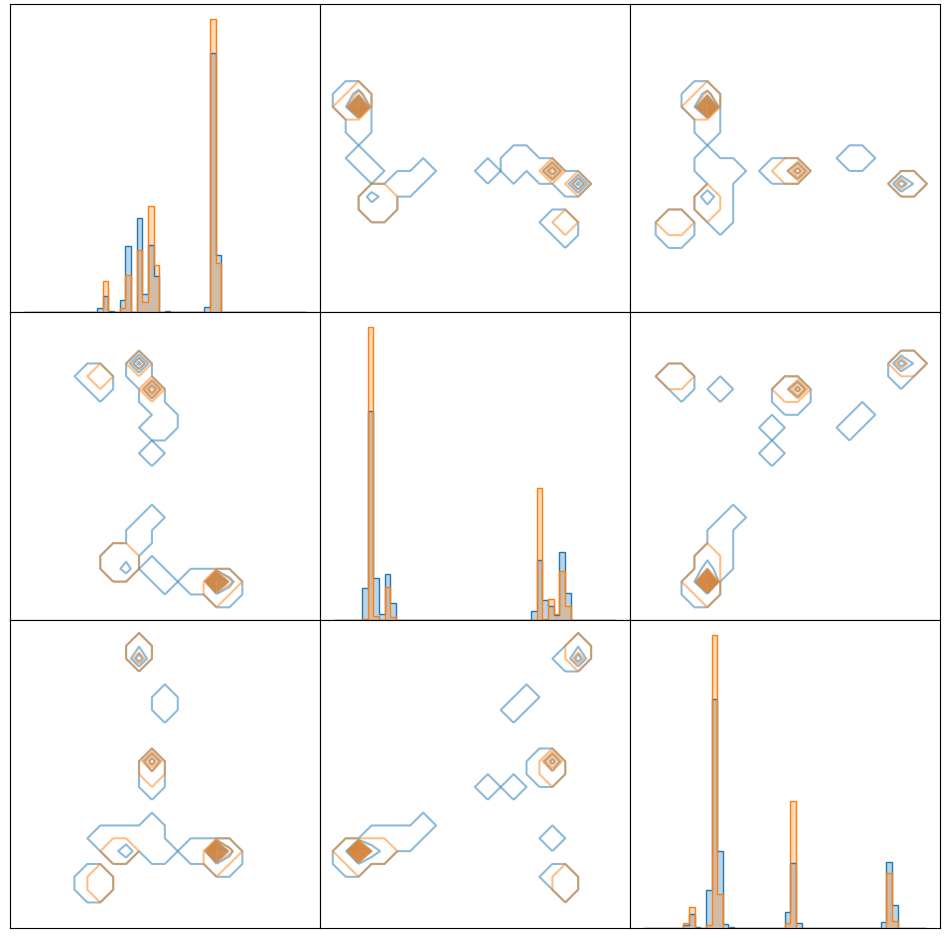}  
  \vspace{-.6cm}
  \caption*{VAE}
\end{subfigure}
\begin{subfigure}{0.49\textwidth}
  \centering
  \includegraphics[width=\textwidth]{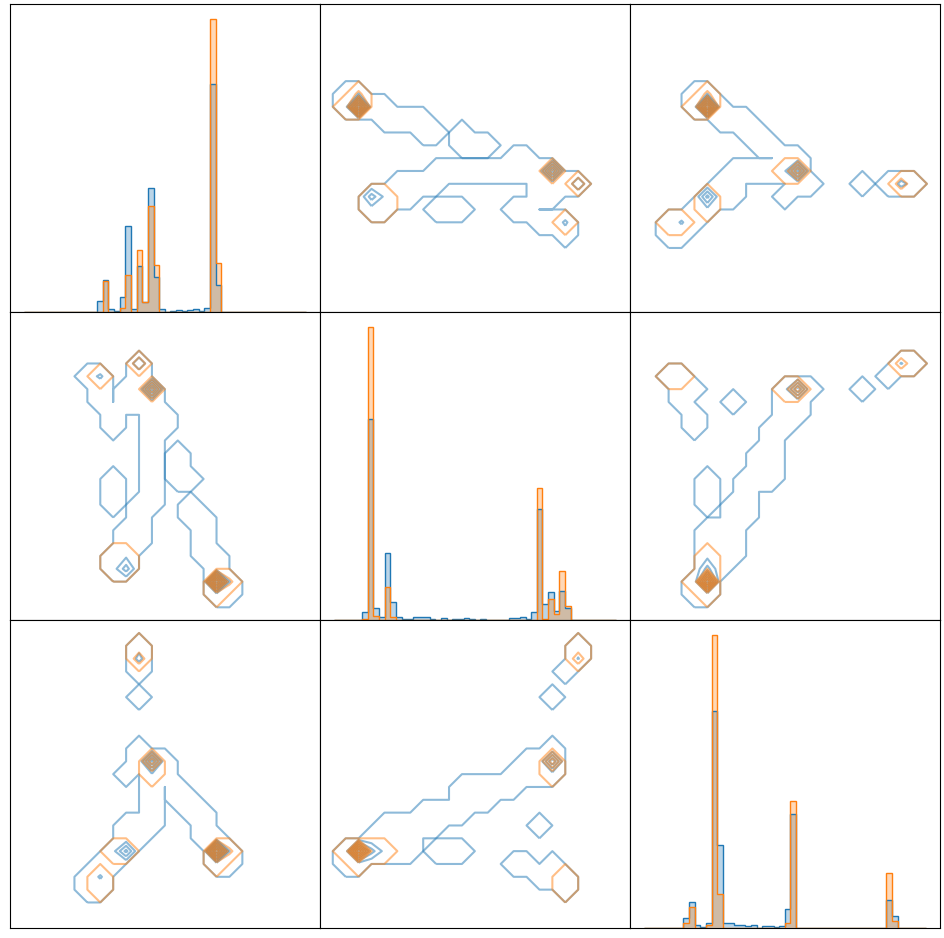}  
  \vspace{-.6cm}
  \caption*{INN + VAE}
\end{subfigure}

\begin{subfigure}{0.49\textwidth}
  \centering
  \includegraphics[width=\textwidth]{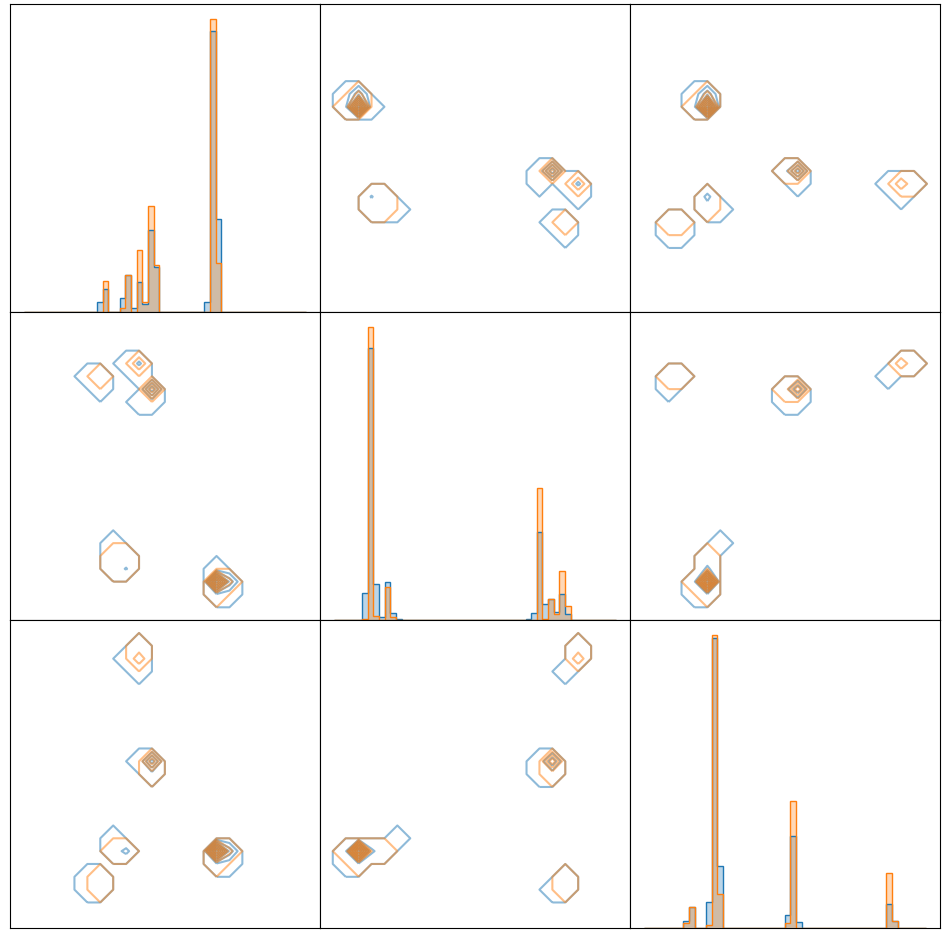}  
  \vspace{-.6cm}
  \caption*{VAE + MALA}
\end{subfigure}
\begin{subfigure}{0.49\textwidth}
  \centering
  \includegraphics[width=\textwidth]{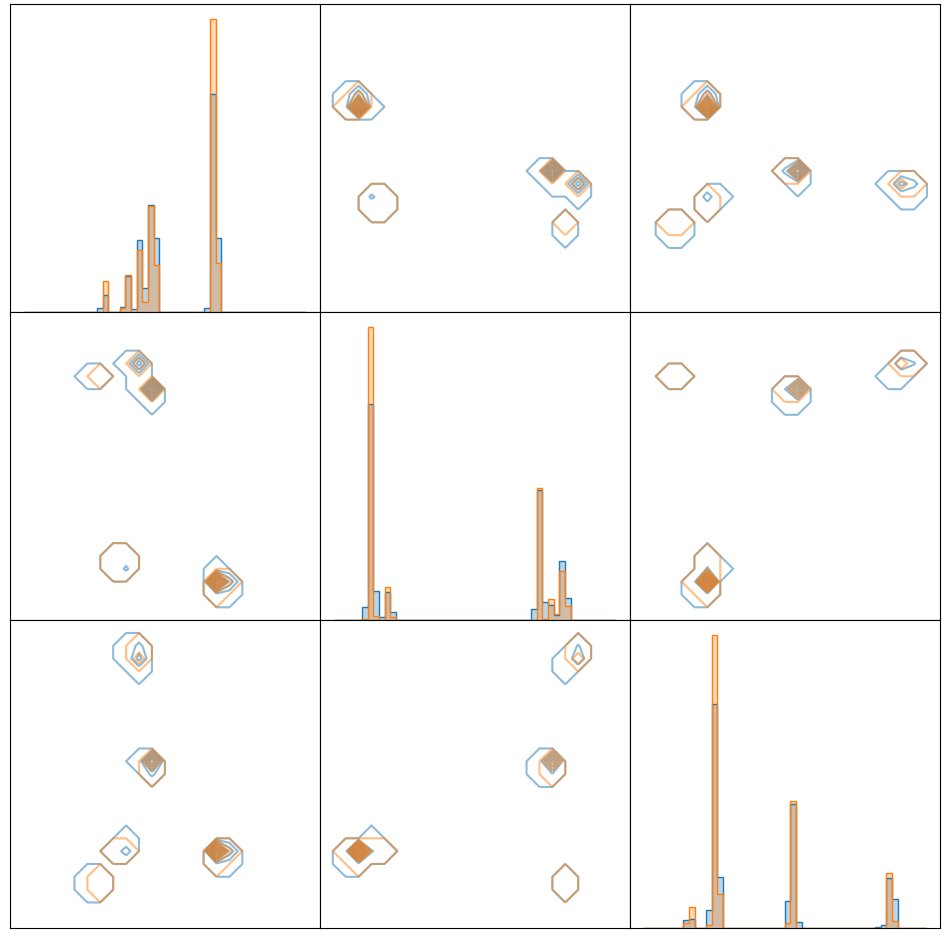}  
  \vspace{-.6cm}
  \caption*{INN+VAE+MALA}
\end{subfigure}
\caption{Histograms and 2-marginals of the first, 
$50$-th and $100$-th marginal of the ground truth posterior distribution (orange) and the posterior reconstructions of the six
different models. 
}
\label{fig:mix_models2}
\end{figure}

 Furthermore, we evaluated the average 
Wasserstein-1 (with respect to the Euclidean loss) distance of the posteriors compared to the ground truth posterior distribution. 

For this, we trained all models 5 times and averaged the Optimal Transport distance with respect to the Euclidean cost function over 100 independently drawn values of $y$. 
The Wasserstein distance was calculated using the Python Optimal Transport package \cite{pot}.

\begin{table}
\scalebox{0.85}{
\begin{tabular}{c|ccccccc}
method&\scriptsize{INN}&\scriptsize{INN+MALA}&\scriptsize{VAE}&\scriptsize{VAE+INN}&\scriptsize{VAE+MALA}&\scriptsize{INN+VAE+MALA}&\scriptsize{FULL FLOW}\\
\hline
$W_1$   &2.12&1.89&1.73&1.55&0.98&0.82&0.92 
\end{tabular}   
}
\caption{Averaged Wasserstein-$1$ distances of the reconstructed posterior distributions using different methods.}
\label{tab_res}
\end{table}

The results are given in Table~\ref{tab_res}.
They indicate that INN+VAE+MALA worked best, but VAE+MALA and FULL FLOW were only slightly worse.
 VAE and VAE+INN performed comparatively.
The conditional INN+MALA were worse, but better than the conditional INN.

The results are depicted in Figure \ref{fig:mix_models} and \ref{fig:mix_models2}.
This numerical evaluation shows in particular that combining different kinds of layers will certainly be helpful when modeling densities: The MALA layer seems to help to anneal to the exact peaks whereas the pure INN models seem to smear those peaks out, as can be seen in the top left and right of the two figures. However, inserting a MALA layer only in the last layer of INNs does not yield very good results, as the MALA layer has trouble with the mixing of different modes, i.e. the mass does not seem to be distributed perfectly.
Furthermore, the variational autoencoder seems to be better at modeling those peaky densities themselves. However note that we did not optimize model parameters that much. For instance, learning proposals or learning the covariance of autoencoders can certainly help to obtain better results. We however followed the rough intuition, that we want to decrease the noise level in the last layer so that we have a chance to model the distribution correctly.

\subsection{Example from Scatterometry}\label{sec:scatterometry}
In this example, we are concerned with a non-destructive technique to determine the structures of photo masks described in more detail by
\cite{heidenreich2015bayesian,scatter} and follows almost the same setup as in \cite{HHS2021}. However here we consider MALA layer as our stochastic layers as well as VAE layers.

The parameters in $x$-space describe the geometry of the photo masks and 
$$Y = F(X) + \eta$$ 
the observed diffraction pattern. The goal is to recover the conditional distribution $P(X| Y= y)$, where the noise model is mixed additive and multiplicative $\eta=a F(X) \eta_1+b\eta_2$, 
where $\eta_1,\eta_2\sim \mathcal{N}(0,I)$ and $a,b>0$ are some constants. 
Then, the conditional distribution $P_{Y|X=x}$ is given by $\mathcal N \left(F(x),(a^2 F(x)^2+b^2) \, I \right)$.
We set $a = 0.2$ and $b = 0.01$. 

The forward operator $F\colon \R^3 \rightarrow \R^{23}$  describes the diffraction of lights, and is physically modelled by solving a PDE, see \cite{scatter}. However, as our method requires gradients with respect to the forward operator $F$, we approximate it using a feedforward neural network.

We choose the prior for $x=(x_1,x_2,x_3)\in\R^3$ by
$$
p_X(x)\coloneqq q(x_1)q(x_2)q(x_3),
$$
where
$$q(x)\coloneqq \begin{cases}\frac{\alpha}{2\alpha+2}\exp(-\alpha(-1-x)),&$for $x<-1,\\
\frac{\alpha}{2\alpha+2},&$for $x\in[-1,1],\\\frac{\alpha}{2\alpha+2}\exp(-\alpha(x-1)),&$for $x>1,\end{cases}
$$
and $\alpha\gg 0$ is some constant, which approximates a uniform distribution for large $\alpha$.
In our numerical experiments, we choose $\alpha=100$.

We repeat the numerical experiment from the previous section with less models. 
In particular, we use the following models with a similar number of parameters (roughly 50000) and trained for the same number of optimizer steps, namely 5000:
\begin{itemize}
\item INN: We use 4 layers with 64 hidden neurons, where each feedforward neural network has two hidden layers.
\item \textbf{INN+MALA}: The same architecture, but with a MALA layer with step size 1e-3 and 3 steps in the last layer. 
\item \textbf{VAE}: Here we use standard ReLU feedforward neural networks with hidden size 64, and 4 layers of them. 
\item \textbf{VAE-MALA}:  Same architecture with a MALA layer with step size 1e-3 and 3 steps.
\item \textbf{Full-Flow}: Here we use two conditional INN layer with intermediate MALA layer steps which anneal to the geometric proposal density and then two conditional VAE layers  with one MALA layer in between and at the end. We use 5 MALA steps per MALA layer with step size 1e-2 and step size 1e-3 in the last layer.
\end{itemize} 
Note that using a MALA layer increases computational effort (roughly by a factor of 2) for the second and  fourth model. The last model uses a MALA layer in between all of the layers. Therefore it is by far the slowest (8 times slower than a plain VAE). We used a batch size of 1600.

Similar to \cite{HHS2021}, we obtain the "ground truth" posterior samples via MCMC, where we apply the Metropolis--Hastings kernel 1000 times with a uniform initialization.

\begin{figure}
\begin{subfigure}{0.49\textwidth}
  \centering
  \includegraphics[width=\textwidth]{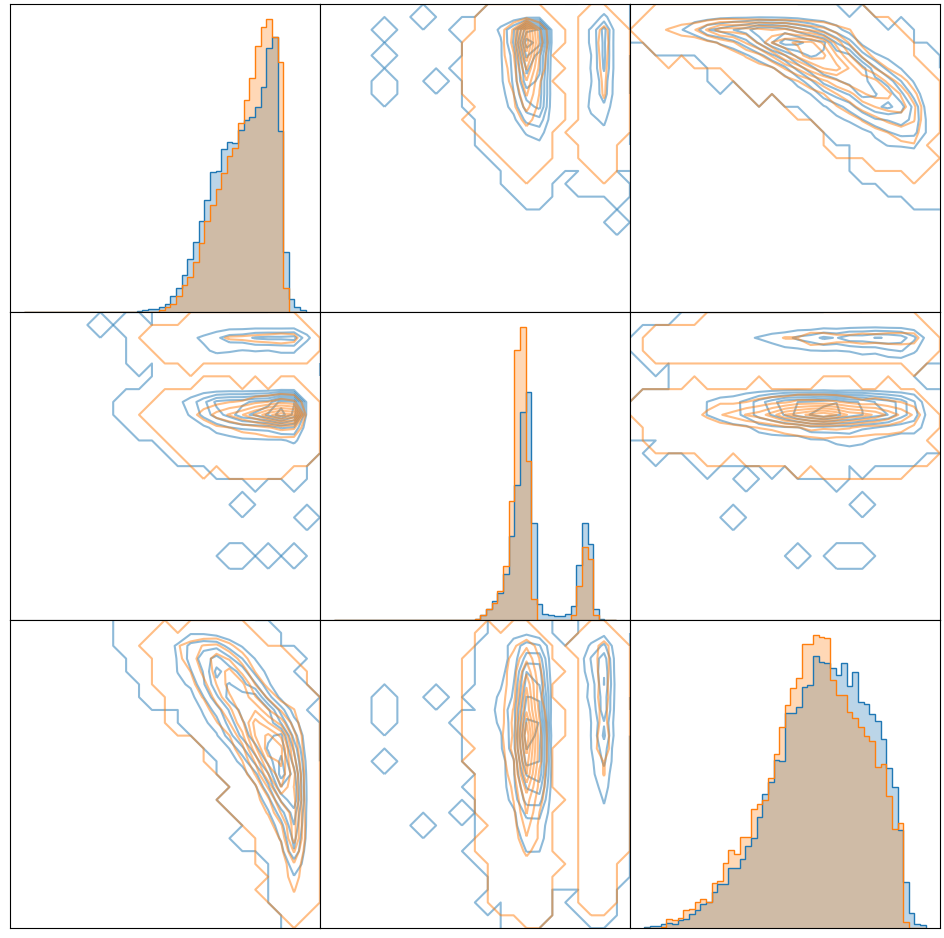}  
  \vspace{-.6cm}
  \caption*{INN}
\end{subfigure}
\begin{subfigure}{0.49\textwidth}
  \centering
  \includegraphics[width=\textwidth]{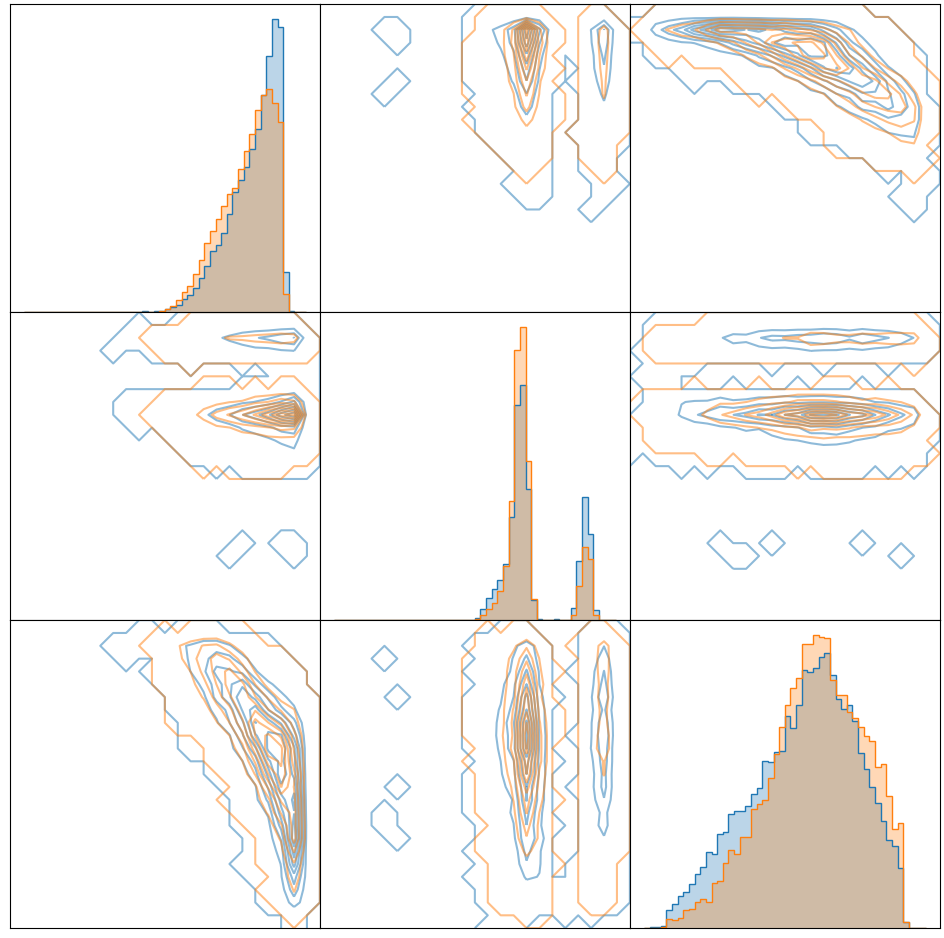}  
  \vspace{-.6cm}
  \caption*{INN + MALA}
\end{subfigure}
	
\begin{subfigure}{0.49\textwidth}
  \centering
  \includegraphics[width=\textwidth]{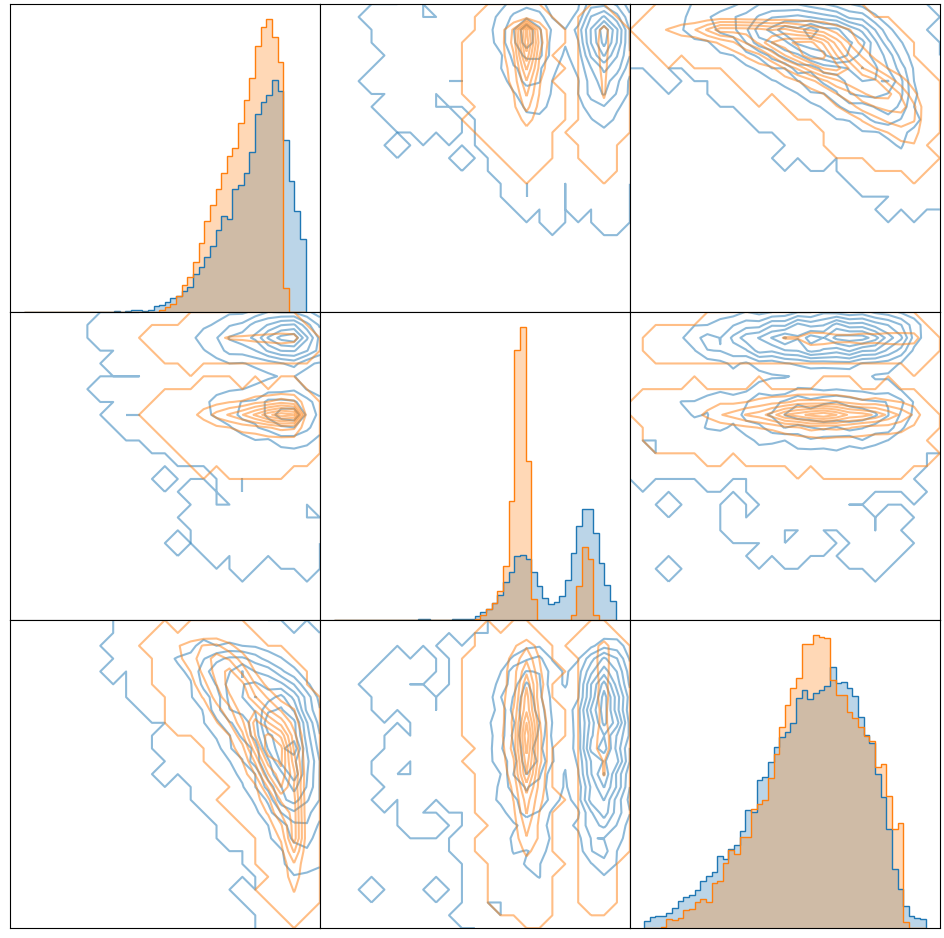}  
  \vspace{-.6cm}
  \caption*{VAE}
\end{subfigure}
\begin{subfigure}{0.49\textwidth}
  \centering
  \includegraphics[width=\textwidth]{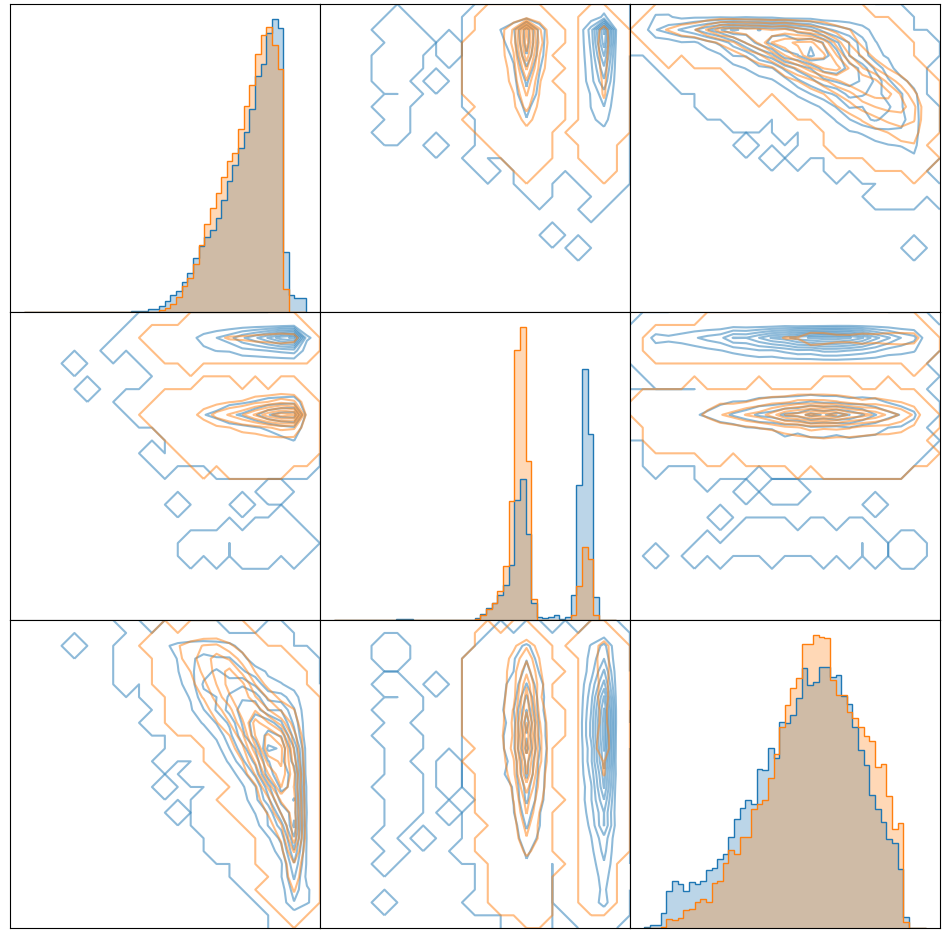}  
  \vspace{-.6cm}
  \caption*{VAE + MALA}
\end{subfigure}
\caption{Histograms of the posterior reconstructions using a different SNF methods in blue and MCMC in orange for 
one sample from $Y$. 
On the diagonal we plot the histograms
of the one-dimensional marginals, on the off-diagonal we plot the distributions of the two 
dimensional marginals.}
\label{fig:scattero}
\end{figure}

\begin{figure}
\begin{subfigure}{0.49\textwidth}
  \centering
  \includegraphics[width=\textwidth]{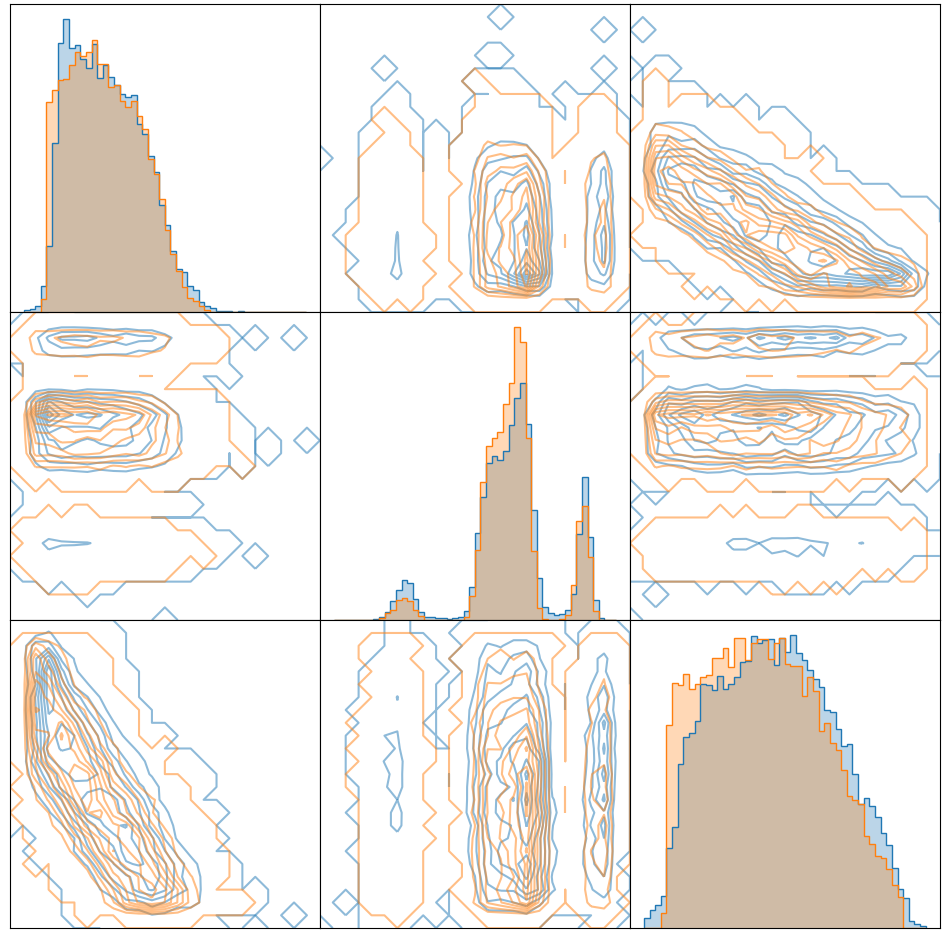}  
  \vspace{-.6cm}
  \caption*{INN}
\end{subfigure}
\begin{subfigure}{0.49\textwidth}
  \centering
  \includegraphics[width=\textwidth]{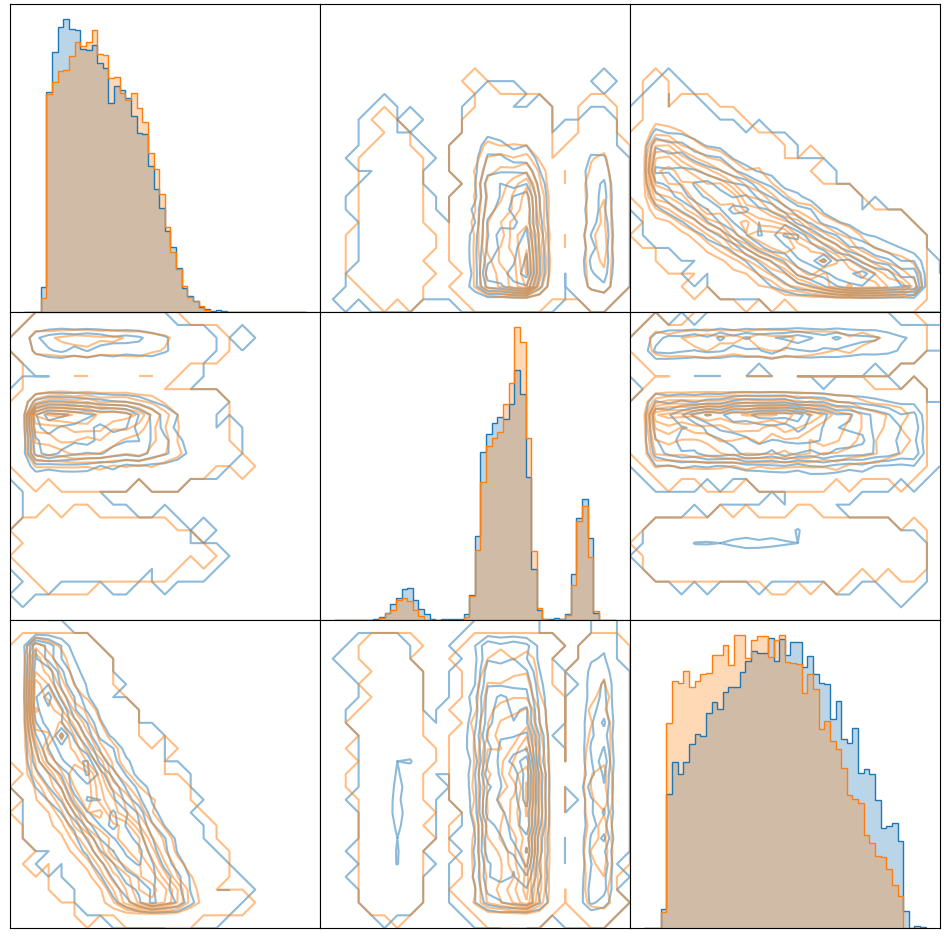}  
  \vspace{-.6cm}
  \caption*{INN + MALA}
\end{subfigure}
	
\begin{subfigure}{0.49\textwidth}
  \centering
  \includegraphics[width=\textwidth]{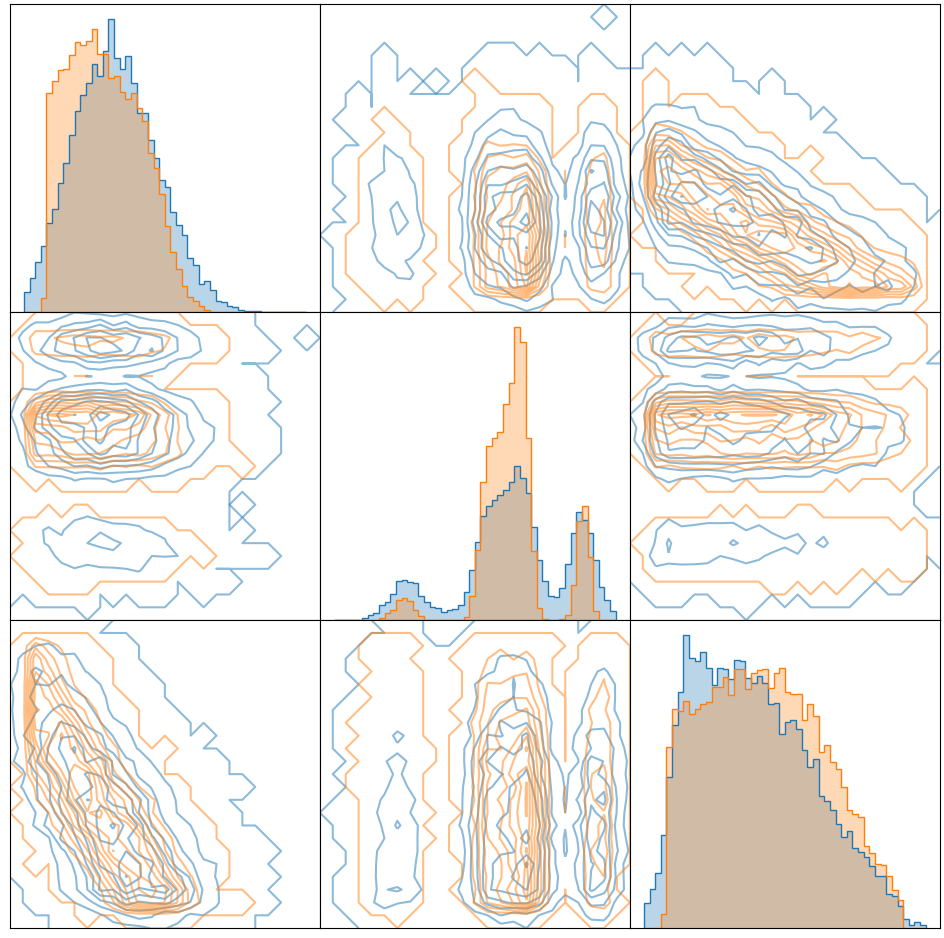}  
  \vspace{-.6cm}
  \caption*{VAE}
\end{subfigure}
\begin{subfigure}{0.49\textwidth}
  \centering
  \includegraphics[width=\textwidth]{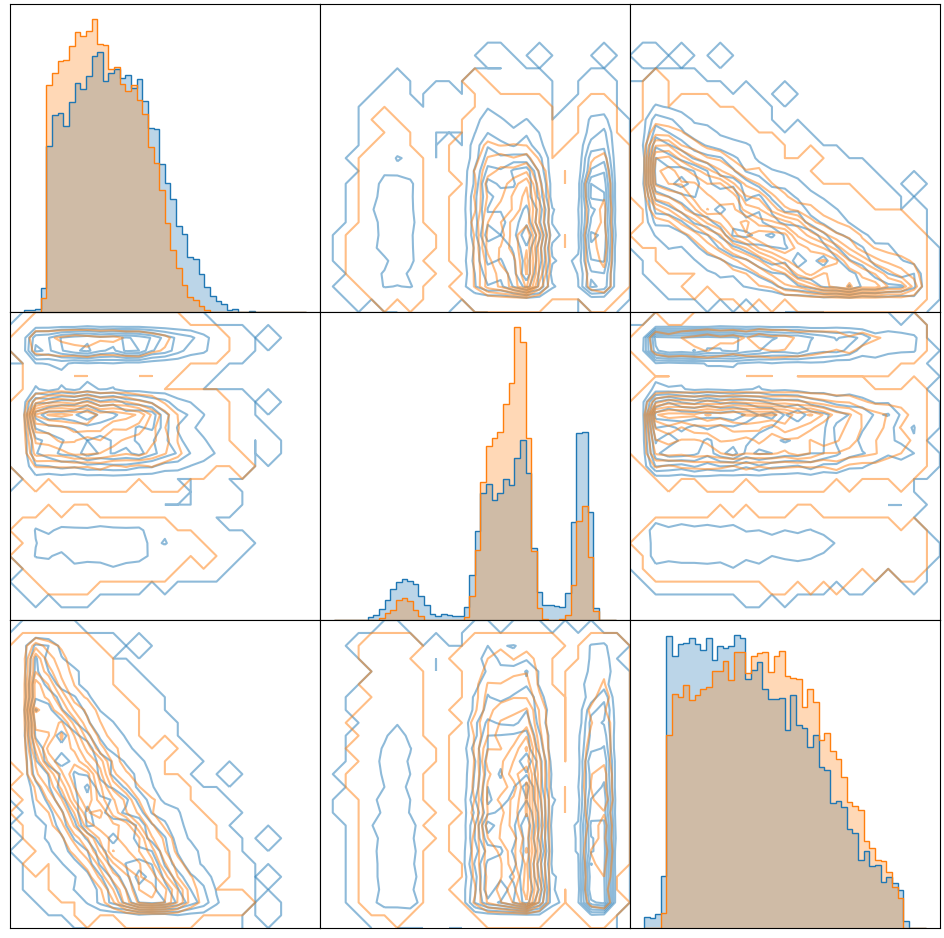}  
  \vspace{-.6cm}
  \caption*{VAE + MALA}
\end{subfigure}
\caption{Histograms of the posterior reconstructions using a different SNF methods in blue and MCMC in orange for 
one sample from $Y$. 
On the diagonal we plot the histograms
of the one-dimensional marginals, on the off-diagonal we plot the distributions of the two 
dimensional marginals.}
\label{fig:scattero2}
\end{figure}

We obtained the following averaged KL-distances $\mathrm{KL}(\mu_\mathrm{MCMC}, \mu_\mathrm{SNF})$ (where we approximated the empirical measures via histograms on $[-1,1]^3$). Results are averaged over 3 training runs, where in each the KL distance is approximated using 20 samples from $P_Y$.
\begin{center}
\begin{tabular}{c|ccccc}
method&\scriptsize{INN}&\scriptsize{INN+MALA}&\scriptsize{VAE}&\scriptsize{VAE+MALA}&\scriptsize{Full Flow}\\
\hline
$KL$  & 0.77&0.63&1.07&0.69&0.60
\end{tabular}                                                                      
\end{center}
In the figures \ref{fig:scattero} and \ref{fig:scattero2} one can see exemplary posterior plots comparing the different methods. 
Basically one can see that the MALA layer improves mass distribution, often removing unnecessary mass from low density regions. Furthermore, the distribution of mass seems best for the INN MALA combination, confirming quantitative results (except for the full flow architecture which is a bit better). 
Note that the VAE architecture could be improved by tuning variances of the respective Markov kernels.

\subsection{Image generation via 2d energy modeling}

\begin{figure}
\begin{center}
  \includegraphics[width=0.9\textwidth]{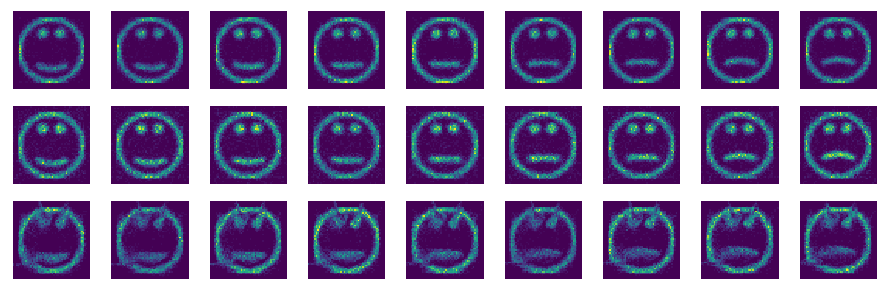}  \\

  \caption{Modeling of smileys as 2d densities with true samples (top row), conditional SNF with plain Gaussian MCMC layer (middle row) and conditional INN (bottom row)}
  \label{cond_smileys}
\end{center}
\end{figure}

\begin{figure}
\begin{center}
  \includegraphics[width=0.9\textwidth]{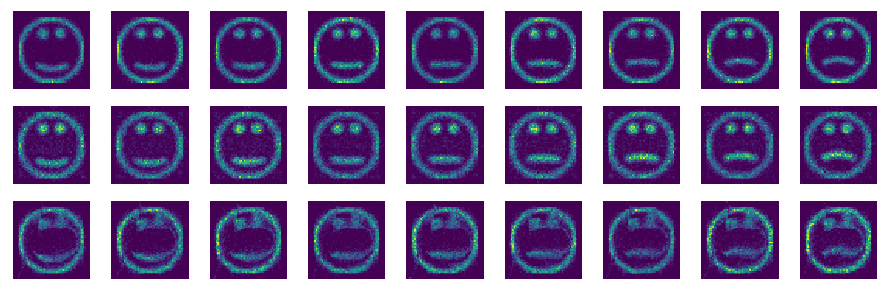}  \\

  \caption{Modeling of smileys as 2d densities with true samples (top row), conditional SNF with MALA layer (middle row) and conditional INN (bottom row)}
  \label{cond_smileys2}
\end{center}
\end{figure}

Finally, we provide a synthetic example for minicking optimal transport between measures.
Suppose we are given two images $\mu_0$ and $\mu_1$
of smileys considered as discrete measures on the image grid
with the associated gray values as weights.
There are many viable ways to interpolate them.
A natural one uses Wasserstein-2 barycenters resulting in the McCann interpolation, see \cite{mccann1997convexity}
$$
\mu_t = ((1-t) \mathrm{id} + t T)_{\#}\mu_0,
$$
where $T$ denotes the optimal transport map between $\mu_0$ and $\mu_1$.
As condition we use a path $(\mu_t)_t$ between two smileys for equispaced values of $t$ in $[0,1]$.
With this at hand, we are finally able to train 
both a conditional SNF as well as a conditional INN on these smileys. 
The results are shown in Figure~\ref{cond_smileys}, 
where the top row shows samples of the conditional SNF 
and the bottom row shows samples of the conditional INN. 
Both have the same number of parameters and were trained 
for roughly the same time. 
However the conditional SNF has more information as we use MCMC layers in between INN layers, 
which need to evaluate the densities. The conditional INN only requires samples for learning. 
That being said, the conditional SNF does a much better job at recovering those concentrated 2d densities, 
as MH layers are able to move the mass from low density regions quite efficiently by jumping. On the other hand, 
the conditional INN clearly captures the shape, but the details get blurred out. 

In Figure \ref{cond_smileys2}, we can see the same experiment repeated with MALA steps instead of Gaussian proposals. Furthermore, we notice that the blurry masses in the INN reconstruction are very dependent on training randomness, as also noted in \cite{WKN2020}.

\newpage
\section{Conclusions and open questions}
In this tutorial paper, we combined several generative models, including MCMC methods, normalizing flows, variational autoencoder and some bits of diffusion models. We want to remark that the way, we put them together in a generalized normalizing flow with a combined loss function is certainly not the only possible way. The idea of stochastic normalizing flows is to match the distributions of forward and backward paths, which also appears in diffusion models \cite{song2021maximum}. It is however not clear, when it is smart to learn a whole path in contrast to just optimizing the final state. 

Moreover, MCMC methods with acceptance-rejection steps involve a non-differentiable step, so that the gradient computation is not unbiased. There is some recent literature which resolve this issue in different ways. The authors of \cite{thin21} derive an unbiased gradient estimate by tracking all the possibilities of acceptance-rejection steps, the authors of \cite{arbel2021annealed, Matthews2022} overcome this issue by optimizing the reverse KL for each distribution individually and the authors of \cite{geffner2021} use uncorrected Hamiltonian steps.

Further, the application for inverse problems in imaging involves the handling of very high-dimensional probability distributions, which causes an extensive computational effort. However, with the increasing resources provided by the hardware development, these problems seem to be tackle-able in future.

\appendix 
\section{Invertible Neural Networks} \label{app:INN}
Invertible neural networks $\mathcal T = \mathcal T(\cdot; \theta)$ with parameters $\theta$, 
which can be used in finite normalizing flows, 
are compositions
\begin{align} \label{eq_net}
  \mathcal T =  T_{L} \circ P_{L} \circ \dots \circ  T_{1} \circ P_{1},
\end{align}
of permutation matrices $P_\ell$ and diffeomorphisms $T_\ell$, $\ell=1,\ldots,L$, i.e., $T_\ell$ are bijections 
and both $T_\ell$ and $T_\ell^{-1}$
are continuously differentiable.
There are mainly two ways to create such invertible mappings, namely as
\begin{itemize}
\item invertible residual neural networks (iResNets) and
\item directly (often coupling based) invertible neural networks (INNs).
\end{itemize}
In the following, we briefly explain these architectures,
where we fix the index and refer just to $T$ instead of $T_\ell$.

Finally, we shortly explain the concept of continuous normalizing flows, which refers to a method of constructing 
an invertible neural network from an ordinary differential equation instead of a concatenation of layers.

\subsection{Invertible ResNets} \label{app:resNets}
Residual Networks were introduced by \cite{HZRS2016}.
Based on the ResNet structure, Behrman et al.~\cite{BGCDJ2019} constructed invertible ResNets and
applied them to various problems, see \cite{CBDJ2019,behrmann2020understanding}.
Here $T$ has the structure
\begin{align} \label{resnet}
T(x) & = x + \Phi(x;\theta) , 
\end{align}
where $\Phi(x;\theta)$ is a Lipschitz continuous subnetwork with Lipschitz constant smaller than 1.
Then given $y = T(x) = x + \Phi(x;\theta)$, the inversion of \eqref{resnet} can be done by considering
the fixed point equation
$$
x = y - \Phi(x;\theta)
$$
and applying the Picard iteration
\begin{align*}
&x^{(0)} = y,\\
&{\mathrm{For}} \quad r=0,1,\ldots\\
&x^{(r+1)}=y - \Phi(x^{(r)};\theta)).
\end{align*}
By Banach's fixed point theorem the sequence $\{x^{(r)}\}_r$ converges to $T^{-1}(y)$. 
In general, much effort needs to be invested in order to control the Lipschitz constant of the subnetworks in the learning process
and several variants were proposed by the above authors.
An alternative would be to restrict the  networks $\Phi(x;\theta)$ to so-called proximal neural networks
which are automatically averaged operators having a Lipschitz constant not larger than one.
Such networks were proposed based on \cite{CP2018}
by \cite{HHNPSS2019,HNS2020}.
But clearly, ,,there is no free lunch'' and the learning process of a proximal neural networks
requires a stochastic gradient descent algorithm on a Stiefel manifold which is more expensive than 
the algorithm in the Euclidean space.
For another approach, we refer to \cite{PRTW2021}.

\subsection{INNs} \label{app:inns}
In INNs, the networks $T$ are directly invertible due to their special structure,
namely either a simple triangular one or a  slightly more sophisticated
block exponential one. We briefly sketch the triangular structure and explain in more detail
the second model, since this is used in our numerical examples. Our numerical examples and code use the FrEIA package \footnote{https://github.com/VLL-HD/FrEIA}.

Real-valued volume-preserving transformations (real NVPs) were introduced by \cite{DSB2017}.
They have the form
$$
  T(\xi_1,\xi_2) 
  := \begin{pmatrix} \xi_1
	\\ 
          \xi_2 \, \mathrm{e}^{s(\xi_1)}   + t(\xi_1)  
					\end{pmatrix},
					\quad 
					\xi = \begin{pmatrix} \xi_1\\\xi_2 \end{pmatrix} \in \mathbb R^d,
$$
where $\xi_1 \in \mathbb R^{d_1}$, $\xi_2 \in \mathbb R^{d_2}$ and $d = d_1 + d_2$.
Here $s, t: \R^{d_2} \to \R^{d_1}$ 
are subnetworks and the product and exponential are taken componentwise. 
Then the inverse map is obviously
$$
  T(x_1,x_2) 
  := \begin{pmatrix} x_1
	\\ 
          \left(x_2 - t(x_1) \right)\, \mathrm{e}^{-s(x_1)}  
					\end{pmatrix},
					\quad x = \begin{pmatrix} x_1\\ x_2 \end{pmatrix} \in \mathbb R^d.
$$
and does  not require an inversion of the feed-forward subnetworks. 
Hence the whole
map $\mathcal T$ is invertible and allows for a fast evaluation of both forward and
inverse map.

In our experiments, we use the slightly sophisticated architecture proposed in \cite{AKRK2019,AFHHSS2021} which has the form
\begin{equation*}  \label{eq_DefBlock}
  T (\xi_1,\xi_2) 
  = (x_1,x_2) 
  \coloneqq \begin{pmatrix}
	\xi_1 \, \mathrm{e}^{s_{2}(\xi_2)} + t_{2}(\xi_2)\\
          \xi_2 \, \mathrm{e}^{s_{1}(x_1)}   + t_{1}(x_1)   
					\end{pmatrix}, 
					\quad
					\xi = \begin{pmatrix} \xi_1
					\\
					\xi_2 \end{pmatrix} \in \mathbb R^d
\end{equation*}
for some splitting $(\xi_1,\xi_2) \in \R^{d}$ with $\xi_i \in \R^{d_i}$, $i=1,2$.  
Here
$s_{2}, t_{2}: \R^{d_2} \to \R^{d_1}$ 
and 
$s_{1}, t_{1}: \R^{d_1} \to \R^{d_2}$ 
are ordinary feed-forward neural networks. 
The parameters $\theta$ of
$\mathcal T(\cdot;\theta)$ are specified by the parameters of these subnetworks.
The inverse of the network layers $T$ is 
\begin{equation*}\label{eq_DefInvBlock}
    T^{-1}(x_1,x_2) 
    = (\xi_1,\xi_2) 
		\coloneqq 
		\begin{pmatrix} 
		\big(x_1 - t_{2}(\xi_2) \big) \,\mathrm{e}^{-s_{2}(\xi_2)}\\ 
             \big(x_2 - t_{1}(x_1)   \big) \,\mathrm{e}^{-s_{1}(x_1)} 
						\end{pmatrix}.
\end{equation*}

In our loss function of the form \eqref{loss_n}, we will need the log-determinant of $\mathcal T$ in \eqref{eq_net}.
Fortunately this can be simply computed by the following  considerations:
since $T_\ell = T_{2,\ell} \circ T_{1,\ell}$ with 
\begin{align*}
    T_{1,\ell}(\xi_1,\xi_2) = (x_1,\xi_2) 
		&\coloneqq
    \left(
		\xi_1\mathrm{e}^{s_{\ell,2}(\xi_2)} + t_{\ell,2} (\xi_2), \xi_2
    \right),
		\\
		T_{2,\ell}(x_1,\xi_2) = (x_1,x_2) 
		&\coloneqq 
		\left(x_1,
    \xi_2\mathrm{e}^{s_{\ell,1}(x_1)} + t_{\ell,1} (x_1) 
		\right), 
\end{align*}
we can readily compute the gradient  
\begin{align*}
		  \nabla T_{1,\ell}(\xi_1,\xi_2) 
			&= 
		  \begin{pmatrix}
		    \mathrm{diag} \left( \mathrm{e}^{s_{\ell,2}(\xi_2)} \right) 
				& \mathrm{diag} \left( \nabla_{\xi_2} 
				\left( \xi_1 \mathrm{e}^{s_{\ell,2}(\xi_2)} + t_{\ell,2} (\xi_2) \right) 
				\right)\\
		    0 & I_{d_2}
		  \end{pmatrix}.
    \end{align*}
    Hence we can use the block structure
		$ \det \nabla T_{1,\ell}(\xi_1,\xi_2) =  \prod_{k=1}^{d_1}
    \mathrm{e}^{\left( s_{\ell,2}(\xi_2)\right)_k} $ 
		and analoguous for 
		$\nabla T_{2,\ell}$. 
		
		By the chain rules and the property of the permutations that the
    Jacobian of $P_\ell$ is  $P_\ell$ with $|\det P_\ell|=1 $, 
		and that $\det (A B) = \det(A) \det(B)$, we conclude 
    \begin{align*}
		  \log( |\det \left(\nabla \mathcal T(\xi) \right)|)
		  = \sum_{\ell = 1}^L \left( \operatorname{sum}\left(s_{\ell,2} \left( (P_\ell \xi^{\ell} )_2 \right)\right) 
		  + \operatorname{sum}\left(s_{\ell,1}\left( (T_{1,\ell} P_\ell \xi^{\ell} )_1 \right) \right)\right),
    \end{align*}
    where $\operatorname{sum}$ denotes the sum of the components of the respective vector,
    $\xi^{1} \coloneqq \xi$ and $\xi^{\ell} = T_{\ell-1} P_{\ell-1}
    \xi^{\ell-1}$, $\ell = 2,\ldots,L$.

\subsection{Continuous Normalizing Flows}
Continuous normalizing flows (CNFs), also called neural ordinary differential equations, were introduced by \cite{CRBD2018} and were further investigated by \cite{GCBSD2018,OFLR2021}, see \cite{RH2021} for an overview.

In contrast to other approaches, the invertible neural network within a CNF does not consist of a concatenation of certain layers, but it is given by the solution of a differential equation.
More precisely, let $f\colon \R^d\times[0,T]\to\R^d$ be the unique solution of
$$
f(\cdot,T)=Id,\quad \partial_t f(x,t)=v \left(f(x,t),t \right),
$$
where $v=v_\theta\colon\R^d\times[0,T]\to\R^d$ is a 
continuously differentiable
neural network with parameters $\theta$ such that for any $\theta$ there exists a constant 
$\lambda_\theta$ such that $v_\theta(\cdot,t)$ is $\lambda_\theta$-Lipschitz continuous for all $t\in[0,T]$.
Then, we define 
$\mathcal T(z)=\mathcal T_\theta(z)\coloneq f(z,0)$.

\begin{remark}[Evaluation of $\mathcal T$]
For arbitrary fixed $z\in\R^d$ (skipping the spatial variable in $f$), 
the value $\mathcal T(z)$ is given by $f(0)$, 
where $f$ is the solution of the ordinary differential equation 
$\dot f(t)=v(f(t),t)$ with the initial value $f(T)=z$.
\end{remark}

Using the theorem of Picard-Lindelöf, it holds that $\mathcal T$ is invertible and that the inverse is given by $\mathcal T^{-1}(x)=g(x,T)$, where
$g\colon\R^d\times[0,T]\to\R^d$ 
is the unique solution of
$$
g(\cdot,0)=Id,\quad \partial_t g(x,t)=v(g(x,t),t).
$$

\paragraph{Derivation of the loss function}
We rely on the loss function 
$$\mathcal L(\theta)=\E_{x\sim P_X}[-\log(p_{\mathcal T_\#P_Z}(x)]$$
in \eqref{soo}.
Evaluating the log-determinant of $\mathcal T$ directly is infeasible, therefore we compute $\mathcal L$ in a different way.
For this, define $p\colon\R^d\times[0,T]\to\R$ by the continuity equation
$$
p(x,T)=p_Z(x),\quad \partial_t p(x,t)+\mathrm{div}\left(p(x,t)v(x,t) \right)=0.
$$
By Lemma 8.6.1 in \cite{Ambrosio}, we know that
$p(\cdot,t)$ is the density of $f(\cdot,t)_\#P_Z$. 
In particular, we have that 
$$p_{\mathcal T_\#P_Z}(x)=p(x,0).$$
Now it holds by the fundamental theorem of calculus that
\begin{align}
\log(p_{\mathcal T_\#P_Z}(x))&=\log(p(x,0))=\log(p(g(x,0),0))\\
&=\log(p(g(x,T),T))-\int_0^T \frac{\dx \log(p(g(x,t),t))}{\dx t} \dx t\\
&=\log(p_Z(g(x,T)))-\underbrace{\int_0^T \frac{\dx \log(p(g(x,t),t))}{\dx t} \dx t}_{\ell(x,T)},
\end{align}
where
$$
\ell(x,t)\coloneq \int_0^t \frac{\dx \log(p(g(x,s),s))}{\dx s} \dx s.
$$
Using the chain rule, we obtain 
\begin{align}
&\quad\frac{\dx }{\dx t}\log(p(g(x,t),t))\\
&=\frac{1}{p(g(x,t),t)}\frac{\dx}{\dx t}p(g(x,t),t)\\
&=\frac{1}{p(g(x,t),t)}\Big(\langle\partial_x p(g(x,t),t),\partial_t g(x,t)\rangle+\partial_t p(g(x,t),t)\Big).
\end{align}
Applying the definition of $g$ for the first term and the continuity equation for the second one, this is equal to
\begin{align*}
&\quad\frac{1}{p(g(x,t),t)}\Big(\langle\partial_x p(g(x,t),t),v(g(x,t),t)\rangle\\
&\quad\qquad-\mathrm{div}(p(g(x,t),t)v(g(x,t),t))\Big)\\
&=\frac{1}{p(g(x,t),t)}\Big(\sum_{i=1}^d \partial_{x_i} p(g(x,t),t)v_i(g(x,t),t)\\
&\quad\qquad-\sum_{i=1}^d \partial_{x_i}(p(\cdot,t)v_i(\cdot,t))(g(x,t))\Big)\\
&=\frac{1}{p(g(x,t),t)}\Big(\sum_{i=1}^d \partial_{x_i} p(g(x,t),t)v_i(g(x,t),t)\\
&\quad\qquad-\partial_{x_i}p(g(x,t),t)v_i(g(x,t),t)-p(g(x,t),t)\partial_{x_i} v_i(g(x,t),t)\Big)\\
&=-\frac{1}{p(g(x,t),t)}\sum_{i=1}^d p(g(x,t),t)\partial_{x_i} v_i(g(x,t),t)\\
&=-\sum_{i=1}^d \partial_{x_i} v_i(g(x,t),t)=-\mathrm{trace}(\nabla v (g(x,t),t)).
\end{align*}
Putting the things together we obtain 
$$
\mathcal L(\theta)=\E_{x\sim P_X}[-\log(p_{\mathcal T_\#P_Z}(x)]=\E_{x\sim P_X}[-\log(p_Z(g(x,T)))+\ell(x,T)]
$$
such that $g$ and $\ell$ solve the differential equations
$$
g(\cdot,0)=Id,\quad \partial_t g(x,t)=v(g(x,t),t)
$$
and
$$
\ell(\cdot,0)=0,\quad\partial_t\ell(x,t)=-\mathrm{trace}(\nabla v (g(x,t),t)).
$$
\begin{remark}
The optimization of $\mathcal L$ requires to differentiate through a solver of the corresponding differential equations, which is in general non-trivial.
For detailed explanations on the optimization of the loss function, we refer to \cite{CRBD2018}.
\end{remark}

\section{Auxiliary Results} \label{app:proofs}
\begin{lemma} \label{lem:reversal}
Let $(X_0,...,X_T)$ be a Markov chain. 
Then 
$(X_T,...,X_0)$ 
is again a Markov chain.
\end{lemma}

\begin{proof}
Using that for $l=T,...,0$ it holds
$$
P_{(X_l,...,X_T)}=P_{(X_l,X_{l+1})}\times P_{X_{l+2}|X_{l+1}
=\cdot}\times\cdots\times P_{X_T|X_{L-1}=\cdot}
$$
we obtain for any measurable rectangle $A_{l+1}\times\cdots\times A_T$ that
\begin{align*}
&\quad P_{(X_T,...,X_l)}(A_L\times\cdots\times A_l)\\
&=\int_{A_{l+1}\times A_l}\int_{A_{l+2}}\cdots\int_{A_{L-1}} P_{X_T|X_{L-1}=x_{L-1}}(A_T)
\dx P_{X_{T-1}|X_{T-2}=x_{T-2}}(x_{T-1})
\\
&\quad \cdots \dx P_{X_{l+2}|X_{l+1}=x_{l+1}}(x_{l+2}) \dx P_{(X_{l+1},X_l)}(x_{l+1},x_l).
\end{align*}
Since $P_{(X_{l+1},X_l)} = 
P_{X_{l+1}}\times P_{X_l|X_{l+1}=\cdot}$ 
and using the definition of 
$P_{X_{l+1}}\times P_{X_l|X_{l+1}=\cdot}$, 
this is equal to
\begin{align*}
&\quad\int_{A_{l+1}\times A_l}\int_{A_{l+2}}\cdots\int_{A_{T-1}} P_{X_T|X_{T-1}
=x_{L-1}}(A_T) \dx P_{X_{T-1}|X_{T-2}=x_{T-2}}(x_{T-1})
\\
&\quad \cdots \dx P_{X_{l+2}|X_{l+1}=x_{l+1}}(x_{l+2}) \dx 
(P_{X_{l+1}}\times P_{X_l|X_{l+1}=\cdot})(x_{l+1},x_l)
\\
&=\int_{A_{l+1}}\int_{A_{l+2}}\cdots\int_{A_{T-1}} 
P_{X_l|X_{l+1}
=
x_{l+1}}(A_l) P_{X_T|X_{T-1}
=
x_{T-1}}(A_T) \dx P_{X_{T-1}|X_{T-2}
=x_{T-2}}(x_{T-1})
\\
&\quad \cdots \dx P_{X_{l+2}|X_{l+1}=x_{l+1}}(x_{l+2})  \dx P_{X_{l+1}}(x_{l+1}).
\end{align*}
By the definition of 
$P_{X_{l+1}}\times P_{X_{l+2}|X_{l+1}=\cdot}\times\cdots\times P_{X_T|X_{T-1}=\cdot}$ 
this is equal to
\begin{align*}
&=\int_{A_{l+1}\times \cdots\times A_T} P_{X_l|X_{l+1}
=
x_{l+1}}(A_l) \dx (P_{X_{l+1}}\times P_{X_{l+2}|X_{l+1}
=
\cdot}\times\cdots\times P_{X_T|X_{T-1}=\cdot})(x_{l+1},...,x_L)
\\
&=\int_{A_{l+1}\times \cdots\times A_T} P_{X_l|X_{l+1}
=x_{l+1}}(A_l) \dx P_{(X_{l+1},...,X_T)}(x_{l+1},...,x_T)\\
&=\int_{A_T\times \cdots\times A_{l+1}} P_{X_l|X_{l+1}
=x_{l+1}}(A_l) \dx P_{(X_L,...,X_{l+1})}(x_T,...,x_{l+1})\\
&=P_{(X_T,...,X_{l+1})}\times P_{X_l|X_{l+1}=\cdot}(A_T\times\cdots\times A_l)
\end{align*}
Summarizing the above equations yields that
$$
P_{(X_T,...,X_l)}(A_T\times\cdots\times A_l)
=
P_{(X_T,...,X_{l+1})}\times P_{X_l|X_{l+1}=\cdot}(A_T\times\cdots\times A_l)
$$
Since the measurable rectangles are a $\cap$-stable generator of the Borel algebra, we obtain that
$$
P_{(X_T,...,X_l)}=P_{(X_T,...,X_{l+1})}\times P_{X_l|X_{l+1}=\cdot}
$$
Using this argument inductively, we obtain that
$$
P_{(X_T,...,X_0)}=P_{X_T}\times P_{X_{T-1}|X_L=\cdot}\times\cdots\times P_{X_0|X_1=\cdot}.
$$
Thus, by the characterization \eqref{eq_path_measure}, $(X_T,...,X_0)$ is a Markov chain.
\end{proof}

The next lemma shows that optimizing paths infact also optimizes marginals. It can be seen as a special case of the data processing inequality, see \cite{Cover2006}, but we give a self-contained proof.
\begin{lemma}\label{lem_KL_marginals}
Let $X,\tilde X\colon\Omega\to \R^{d_1}$ and $Y,\tilde Y\colon\Omega\to\R^{d_2}$ 
be random variables such that $X$ and $Y$ as well as
$\tilde X$ and $\tilde Y$ have joint strictly positive densities $p_{X,Y}$ and $p_{\tilde X,\tilde Y}$.
Then it holds
$$
\mathrm{KL}(p_X,p_{\tilde X})\leq\mathrm{KL}(p_{X,Y},p_{\tilde X,\tilde Y}).
$$
\end{lemma}

\begin{proof}
Using the law of total probability, we obtain
\begin{align}
\mathrm{KL}(p_{X,Y},p_{\tilde X,\tilde Y})
&=\int_{\R^{d_2}} \int_{\R^{d_1}} p_{X,Y}(x,y) \log(\tfrac{p_{X,Y}(x,y)}{p_{\tilde X,\tilde Y}(x,y)}) \dx x \dx y\\
&=\int_{\R^{d_2}} \int_{\R^{d_1}} p_X(x)p_{Y|X=x}(y) \big(\log(\tfrac{p_X(x)}{p_{\tilde X}(x)})
+\log(\tfrac{p_{Y|X=x}(y)}{p_{\tilde Y|\tilde X=x}(y)})\big) \dx x \dx y\\
&=
\int_{\R^{d_1}} p_X(x) \Big( \int_{\R^{d_2}} p_{Y|X=x}(y)\log(\tfrac{p_X(x)}{p_{\tilde X}(x)}) \dx y \\
& \quad 
+\int_{\R^{d_2}} p_{Y|X=x}(y)\log(\tfrac{p_{Y|X=x}(y)}{p_{\tilde Y|\tilde X=x}(y)})\dx y \Big) \dx x\\
&=
\int_{\R^{d_1}} p_X(x)\underbrace{\int_{\R^{d_2}} p_{Y|X=x}(y)dy}_{=1}\log(\tfrac{p_X(x)}{p_{\tilde X}(x)}) \dx x\\
&
\quad
+\int_{\R^{d_1}} p_X(x)\int_{\R^{d_2}} p_{Y|X=x}(y)\log(\tfrac{p_{Y|X=x}(y)}{p_{\tilde Y|\tilde X=x}(y)}) \dx y \dx x\\
&=\mathrm{KL}(p_X,p_{\tilde X}) + \int_{\R^{d_1}} p_X(x) \, \mathrm{KL}(p_{Y|X=x},p_{\tilde Y|\tilde X=x}) \dx x\\
&\geq \mathrm{KL}(p_X,p_{\tilde X}).
\end{align}
This proves the claim.
\end{proof}

The following lemma gives a derivation of the Markov kernel
for a general form of the Metropolis-Hastings algorithm. 
Let $X_t'$ be a random variable and $U\sim\mathcal U_{[0,1]}$ 
such that 
$\left( \boldsymbol{\sigma}(X_t'),\boldsymbol{\sigma}(U),\boldsymbol{\sigma} \left( \cup_{s\le t-2} \boldsymbol{\sigma}( X_s) \right) \right)$ 
are independent. 
Further, we assume that the joint distribution $P_{X_{t-1},X_t'}$ is given by
$$
P_{X_{t-1},X_t'}=P_{X_{t-1}}\times Q_t
$$
for some appropriately chosen Markov kernel $Q_t\colon\R^d\times\mathcal B(\R^d)\to[0,1]$,
where $Q_t(x,\cdot)$ is assumed to have the strictly positive probability density function $q_t(\cdot|x)$.
We considered the special cases
\begin{itemize}
\item MH layer:
$$
    Q_t(x,\cdot)=\mathcal N(x,\sigma^2 I),\qquad q(\cdot|x)=\mathcal N(\cdot|x,\sigma^2 I).
    $$
\item MALA layer:		
$$
    Q_t(x,\cdot)=\mathcal N(x-a_1\nabla u_t(x),a_2^2 I), \qquad q(\cdot|x)=\mathcal N(\cdot|x-a_1\nabla u_t(x),a_2^2 I).
    $$
\end{itemize}

Then we have the following lemma, see also \cite{T1998}.

\begin{lemma} \label{MH kernel}
Let $X_t'$ be a random variable such that $(X_0,...,X_{t-1},X_t')$ is a Markov chain with
Markov kernel 
$Q_t\coloneqq P_{X_t'|X_{t-1}}\colon\R^d\times\mathcal B(\R^d)\to[0,1]$. Assume that
$P_{X_t'|X_{t-1} = x}$
admits a density $q_t(\cdot|x)$. Set
$$
\alpha_t(x, y ) \coloneqq \min \big\{ 1,\frac{p_t(y) q_t(y|x)}{p_t(x) q_t(x|y)} \big\}.
$$
Further, let $U$ be uniformly distributed on $[0,1]$ and independent of $(X_0,...,X_{t-1},X_t')$.
Then, for $X_t$ defined by
\begin{align}
X_t 
&\coloneqq
1_{[U,1]} \left( \alpha_t( X_{t-1},X_t') \right) \, X_t'
+
1_{[0,U[}  \left( \alpha_t( X_{t-1}, X_t' ) \right) \, X_{t-1},
\end{align} 
the transition kernel $P_{X_t|X_{t-1}}$ is given by
$$
\mathcal K_t(x,A) = \int_A q_t(y|x) \alpha_t (x,y) \dx y + \delta_x(A) \int_{\mathbb R^d} q_t(y|x) \left(1-\alpha_t(x,y) \right)\dx y.
$$
\end{lemma}

\begin{proof}
For any measurable sets $A,B\in\mathcal B(\R^d)$, it holds
\begin{align*}
P_{X_{t-1},X_t}(A\times B)&=\int_\Omega 1_{\{X_{t-1}\in A\}}(\omega) 1_{\{X_t\in B\}}(\omega)\dx P(\omega)\\
&=\int_\Omega 1_{\{U<\alpha_t(X_{t-1},X_t')\}}(\omega)1_{\{X_{t-1}\in A\}}(\omega) 1_{\{X_t\in B\}}(\omega)\dx P(\omega)\\
&+\int_\Omega 1_{\{U\geq\alpha_t(X_{t-1},X_t')\}}(\omega)1_{\{X_{t-1}\in A\}}(\omega) 1_{\{X_t\in B\}}(\omega)\dx P(\omega)
\end{align*}
Since it holds $X_t=X_t'$ on $\{U<\alpha_t(X_{t-1},X_t')\}$ and $X_t=X_{t-1}$ on $\{U\geq\alpha_t(X_{t-1},X_t')\}$, this is equal to
\begin{align*}
&\quad\int_\Omega 1_{\{U<\alpha_t(X_{t-1},X_t')\}}(\omega)1_{\{X_{t-1}\in A\}}(\omega) 1_{\{X_t'\in B\}}(\omega)\dx P(\omega)\\
&+\int_\Omega 1_{\{U\geq\alpha_t(X_{t-1},X_t')\}}(\omega)1_{\{X_{t-1}\in A\cap B\}}(\omega)\dx P(\omega)\\
&=\int_{A\times B\times[0,1]} 1_{[0,\alpha_t(x_{t-1},x_t')]}(u)\dx P_{X_{t-1},X_t',U}(x_{t-1},x_t',u)\\
&+\int_{(A\cap B)\times\R^{d}\times[0,1]} 1_{[\alpha_t(x_{t-1},x_t'),1]}(u)\dx P_{X_{t-1},X_t',U}(x_{t-1},x_t',u).
\end{align*}
As $U$ is independent of $(X_{t-1},X_t')$, this can be rewritten as
\begin{align*}
&\quad\int_{A\times B} \int_{[0,1]}1_{[0,\alpha_t(x_{t-1},x_t')]}(u)\dx P_U(u)\dx P_{X_{t-1},X_t'}(x_{t-1},x_t')\\
&+\int_{(A\cap B)\times\R^{d}}\int_{[0,1]} 1_{[\alpha_t(x_{t-1},x_t'),1]}(u)\dx P_U(u)\dx P_{X_{t-1},X_t'}(x_{t-1},x_t')\\
&=\int_{A\times B} P_U([0,\alpha_t(x_{t-1},x_t')]) \dx P_{X_{t-1},X_t'}(x_{t-1},x_t')\\
&+\int_{(A\cap B)\times\R^{d}}P_U([\alpha_t(x_{t-1},x_t'),1])\dx P_{X_{t-1},X_t'}(x_{t-1},x_t').
\end{align*}
Since $P_U$ is the uniform distribution on $[0,1]$, the above formula becomes
\begin{align*}
P_{X_{t-1},X_t}(A\times B)
&=
\int_{A\times B} \alpha_t(x_{t-1},x_t') \dx P_{X_{t-1},X_t'}(x_{t-1},x_t')\\
&\quad +\int_{(A\cap B)\times\R^{d}}(1-\alpha_t(x_{t-1},x_t'))\dx P_{X_{t-1},X_t'}(x_{t-1},x_t')\\
&=\int_{A\times B} \alpha_t(x_{t-1},x_t')p_{X_{t-1},X_t'}(x_{t-1},x_t') \dx (x_{t-1},x_t')\\
&\quad +\int_{(A\cap B)\times\R^{d}}(1-\alpha_t(x_{t-1},x_t'))p_{X_{t-1},X_t'}(x_{t-1},x_t')\dx (x_{t-1},x_t').
\end{align*}
Further, by definition, we have that $P_{X_{t-1},X_t'}=P_{X_{t-1}}\times Q_t$ such that $P_{X_{t-1},X_t'}$ has the density
$$
p_{X_{t-1},X_t'}(x,y)=q_t(x|y)p_{X_{t-1}}(y).
$$
Thus, we get
\begin{align*}
P_{X_{t-1},X_t}(A\times B)
&=\int_{A}\int_B \alpha_t(x_{t-1},x_t')q_t(x_t'|x_{t-1}) \dx x_t' p_{X_{t-1}}(x_{t-1})\dx x_{t-1}\\
&\quad +\int_{A\cap B}\int_{\R^{d}}(1-\alpha_t(x_{t-1},x_t'))q_t(x_t'|x_{t-1})\dx x_t'p_{X_{t-1}}(x_{t-1})\dx x_{t-1}\\
&=\int_{A}\int_B \alpha_t(x_{t-1},x_t')q_t(x_t'|x_{t-1}) \dx x_t' \dx P_{X_{t-1}}(x_{t-1})\\
&\quad +\int_{A}\delta_{x_{t-1}}(B)\int_{\R^{d}}(1-\alpha_t(x_{t-1},x_t'))q_t(x_t'|x_{t-1})\dx x_t'\dx P_{X_{t-1}}(x_{t-1})\\
&=
\int_{A} \Big(\int_B \alpha_t(x_{t-1},x_t')q_t(x_t'|x_{t-1}) \dx x_t' \\
&\quad +\delta_{x_{t-1}}(B)\int_{\R^{d}}(1-\alpha_t(x_{t-1},x_t'))q_t(x_t'|x_{t-1})\dx x_t' \Big) \dx P_{X_{t-1}}(x_{t-1})\\
&=\int_{A}\mathcal K_t(x_{t-1},B)\dx P_{X_{t-1}}(x_{t-1}).
\end{align*}
In summary, we obtain 
$$
P_{X_{t-1},X_t}(A\times B)=\int_{A}\mathcal K_t (x_{t-1},B)\dx P_{X_{t-1}}(x_{t-1}).
$$
As the measurable rectangles are a $\cap$-stable generator of $\mathcal B(\R^d\times\R^d)$, 
this yields that $P_{X_{t-1},X_t}=P_{X_{t-1}}\times \mathcal K_t$ such that $P_{X_t|X_{t-1}}=\mathcal K_t$.
\end{proof}

\bibliography{ref}

\section*{Acknowledgements}
The funding by the German Research Foundation (DFG) within 
 the projects STE 571/16-1 and 
within the project of the DFG-SPP 2298 ,,Theoretical Foundations of Deep Learning'' 
is gratefully acknowledged.
Many thanks to S. Heidenreich from the Physikalisch-Technische Bundesanstalt (PTB) for
providing the scatterometry data which we used for training the forward model 
and for fruitful discussions on the corresponding example. 
We would like to thank A. Houdard for generating the image on the bottom of Figure~\ref{fig:images}.
Many thanks to V. Stein for proofreading.

\end{document}